\documentclass{article}

\PassOptionsToPackage{numbers, compress}{natbib}

 \usepackage[preprint]{neurips_2026}

\usepackage[utf8]{inputenc} 
\usepackage[T1]{fontenc}    
\usepackage{url}            
\usepackage{booktabs}       
\usepackage{amsfonts}       
\usepackage{nicefrac}       
\usepackage{microtype}      
\usepackage{xcolor}         

\usepackage{amsmath}
\usepackage{amssymb}

\usepackage{multirow}
\usepackage{makecell}
\usepackage{graphicx}
\usepackage{graphicx, subcaption}

\usepackage{wrapfig}

\definecolor{citecolor}{RGB}{48,111,186}
\usepackage[pagebackref,breaklinks=true,colorlinks,citecolor=blue,urlcolor=blue,linkcolor=blue,bookmarks=false]{hyperref}
\AtEndPreamble{
    \usepackage[capitalize]{cleveref}
    \crefname{section}{Sec.}{Secs.}
    \Crefname{section}{Section}{Sections}
    \crefname{table}{Tab.}{Tabs.}
    \Crefname{table}{Table}{Tables}
}
\usepackage[table]{xcolor}
\usepackage{multicol}
\usepackage{booktabs}
\usepackage{colortbl}

\usepackage{algorithm}
\usepackage{algorithmic}

\title{Res\(^2\)CLIP: Few-Shot Generalist Anomaly Detection with Residual-to-Residual Alignment}

\author{%
  \textbf{Xinyue Liu}$^{1}$, \textbf{Jianyuan Wang}$^{2*}$, \textbf{Biao Leng}$^{1}$, 
  \textbf{Shuo Zhang}$^{3}$ \\
  $^1$Beihang University,
  $^2$University of Science and Technology Beijing, 
  $^3$Beijing Jiaotong University
}

\begin{document}

\maketitle

\begin{abstract}

Few-shot Generalist Anomaly Detection requires models to generalize to novel categories without retraining, posing significant challenges in real-world scenarios with scarce samples and rapidly changing categories. Existing CLIP-based methods face two major challenges: coarse-grained unified text prompts struggle to adapt to fine-grained foreground-background differences, causing cross-granularity mismatch; and fine-tuning on auxiliary datasets disrupts CLIP's inherent open-world generalization due to domain shift, leading to cross-category generalization degradation. To address these, we propose to shift multimodal alignment entirely into a unified residual space, where residual representations naturally eliminate fine-grained normal feature differences across regions and class-specific biases, simultaneously resolving both problems. Based on this insight, Res\(^2\)CLIP, the first residual-to-residual alignment framework that symmetrically bridges visual and text modalities within CLIP's residual space, is designed. The framework is developed from a residual perspective into three branches: a text prompt-based branch, a visual prompt-based branch, and a novel residual-to-residual alignment branch.
All learnable optimizations are constrained within the residual domain, and the residual alignment optimization objectives are designed to force the model to focus on relative anomaly deviations rather than optimizing class-specific features. Experiments on multiple datasets demonstrate the effectiveness of our architecture.
The code is available at \url{https://github.com/hito2448/Res2CLIP}.

\end{abstract}
\section{Introduction}
\label{sec:intro}

Anomaly Detection (AD) is critical for real-world applications such as industrial inspection, but sample scarcity, frequent product changes, and strict privacy constraints often make retraining infeasible. This drives few-shot Generalist Anomaly Detection (GAD), which requires a model trained only on auxiliary data to immediately generalize to  novel data domains. 
Recently, anomaly detection methods built on vision-language models such as CLIP  ~\cite{radford2021learning} have become mainstream for few-shot GAD ~\cite{jeong2023winclip, zhu2024toward}, typically adopting a dual-branch design with text prompts for semantic scoring and visual prompts for local similarity matching.

\begin{figure}
    \centering
    \includegraphics[width=\linewidth]{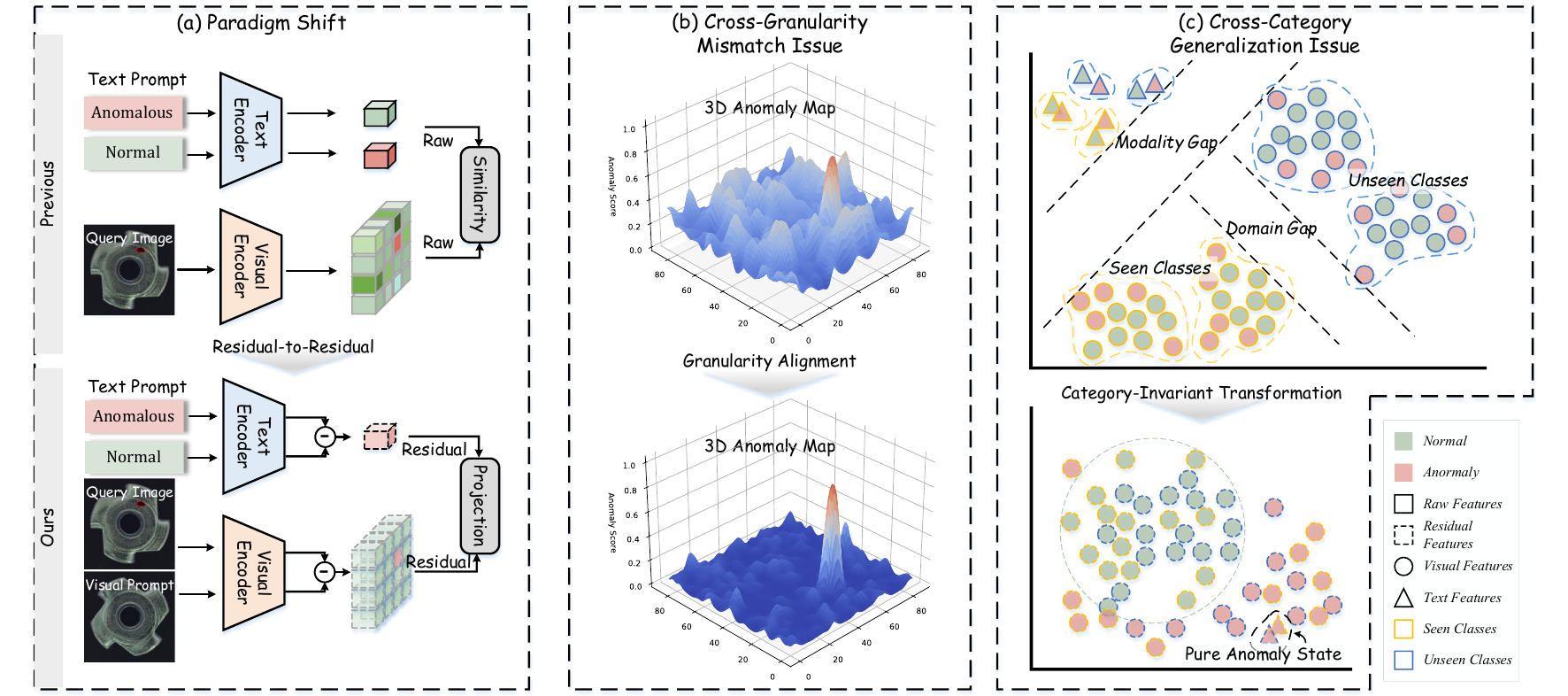}
    \caption{ Motivation and comparison with prior work. Existing CLIP-based methods suffer from cross-granularity mismatch and cross-category generalization degradation. Res\(^2\)CLIP mitigates both via residual-to-residual alignment, which enables precise granularity alignment and transforms features into category-agnostic representations.
    }
    \label{fig:motivation}
\end{figure}

However, directly relying on text-visual alignment for anomaly detection within the original CLIP space presents critical limitations:
(1) \textbf{Cross-Granularity Mismatch}: 
Foreground objects and background regions exhibit vastly different features. Using a unified, coarse text prompt to measure these fine-grained local regions causes feature mismatch, often misclassifying normal backgrounds as anomalies as in \cref{fig:motivation} (b).
(2) \textbf{Cross-Category Generalization Degradation}: 
To adapt to specific tasks, existing methods fine-tune on seen categories to optimize the alignment between text and visual features, such as by introducing learnable prompts or modality-specific feature adapters. 
However, as in ~\cref{fig:motivation} (c), features from each category cluster together in the original CLIP space. Fine-tuning within this space forces the model to overfit to seen training categories, which weakens CLIP's intrinsic open-world generalization and degrades cross-category performance on unseen data.

To address these issues, our intuition is to shift the multimodal alignment into a unified \textbf{\textit{residual domain}}. 
Specifically, we construct residual representations for both modalities: the textual residual is the difference between ``anomalous'' and ``normal'' text prompts, while the visual residual subtracts corresponding normal features retrieved among few-shot visual prompts from the query features.
These representations not only eliminate foreground-background granularity discrepancies to resolve the cross-granularity mismatch, but also smooths out specific category information to alleviate the generalization problem.

Consequently, we propose Res\(^2\)CLIP, a novel multimodal alignment framework that symmetrically bridges visual and text modalities through residual features. 
We first reconstruct the previous dual-branch paradigm from a residual perspective, transforming the original visual and text feature extraction into residual feature extraction. Building upon this, we further propose a text-visual residual-to-residual alignment mechanism within CLIP's residual space. Notably, all fine-tuning optimizations are conducted entirely within this residual domain, which forces the model to focus on learning relative deviations rather than memorizing class-specific features. This effectively alleviates overfitting, achieving precise localization while preserving CLIP's generalization. The main contributions of this paper are summarized as follows:
\begin{itemize}
        \item 
        We unify the alignment of CLIP's text and visual features into the residual domain for the few-shot GAD task, thereby alleviating the granularity and generalization issues.
        \item A three-branch few-shot GAD architecture is designed from a residual perspective, which includes the widely recognized text and visual branches, and our developed residual branch.
        \item All optimization operations are shifted to the residual domain, and suitable loss designs are formulated for the alignment within this space.
        \item Experimental verification on multiple datasets demonstrate the effectiveness and  superior localization capability of our architecture.
\end{itemize}

\section{Related Work}

\noindent
\textbf{Few-Shot Anomaly Detection.}
Unsupervised anomaly detection methods rely on abundant normal samples to
learn reconstruction patterns~\cite{you2022unified, zhang2023unsupervised},
feature distributions~\cite{gudovskiy2022cflow, zhou2024msflow}, or feature
modeling~\cite{deng2022anomaly, guo2023template, gu2023remembering,
guo2024recontrast, liu2025unlocking}, but their performance degrades sharply
in few-shot settings.
Few-shot anomaly detection methods address this by discriminating and
localizing anomalies from only a handful of normal samples per class.
Training-free approaches such as PaDiM~\cite{defard2021padim} and
PatchCore~\cite{roth2022towards} construct memory banks or estimate
distributions from the available samples and detect anomalies via
nearest-neighbor or Mahalanobis distance, but do not learn transferable
detection capabilities. Optimization-based approaches including
RegAD~\cite{huang2022registration}, FastRecon~\cite{fang2023fastrecon},
PromptAD~\cite{li2024promptad}, IIPAD~\cite{lv2025one}, and
FoundAD~\cite{zhai2025foundation} fine-tune registration networks, prompts,
or manifold projections, but still require per-class re-modeling.
To remove this per-class dependency, the generalist paradigm trains a unified
model on auxiliary categories and applies it directly to novel ones without
adaptation. WinCLIP~\cite{jeong2023winclip} and APRIL-GAN~\cite{chen2023zero}
inject CLIP's language priors into few-shot detection, and
InCTRL~\cite{zhu2024toward} formally defines the generalist task by combining
text prompts with query-support residuals. Subsequent methods explore
visual and textual perspectives in different ways:
ResAD~\cite{yao2024resad} leverages residual features within the visual
feature space to eliminate domain shifts;
MetaUAS~\cite{gao2024metauas} unifies anomaly segmentation into change
segmentation; DictAS~\cite{qu2025dictas} reformulates detection as
dictionary lookup; AdaptCLIP~\cite{gao2026adaptclip} inserts adaptive
adapters for zero-/few-shot generalization; and
ReMP-AD~\cite{ma2025remp} fuses retrieval-augmented multimodal prompts.
Among these, ResAD is closest to our work in adopting a residual
representation, but operates entirely within the visual modality. In
contrast, we extend the residual paradigm to a symmetric multimodal
alignment between visual and text residuals.

\noindent
\textbf{CLIP-based Anomaly Detection}
Recently, researchers have introduced pre-trained vision-language models such as CLIP~\cite{radford2021learning} into anomaly detection, leveraging their massive pre-trained knowledge and strong vision-text alignment capability. 
WinCLIP~\cite{jeong2023winclip} pioneered hand-crafted normal/abnormal
prompt templates for language-guided zero-/few-shot detection.
AnomalyCLIP~\cite{zhou2023anomalyclip} learns object-agnostic prompts to
capture generic normality and abnormality patterns.
AdaCLIP~\cite{cao2024adaclip} introduces static-dynamic hybrid prompts for
test-time adaptation. AA-CLIP~\cite{ma2025aa} constructs anomaly-aware
textual anchors and aligns patch-level visual features with these anchors
through a two-stage strategy. Bayes-PFL~\cite{qu2025bayesian} models the
prompt space as a learnable distribution from a Bayesian perspective.
VCP-CLIP~\cite{qu2024vcp} injects global visual context into prompts to
eliminate product-specific dependencies, and MRAD~\cite{xu2026mrad}
constructs a memory bank for direct anomaly score retrieval.
These methods all perform alignment within CLIP's original feature space,
where unified text embeddings struggle to capture fine-grained
foreground-background differences and domain-specific optimization risks
overfitting to seen classes. Our work instead shifts multimodal alignment
into the residual domain.

\section{Method} \label{sec:method}

\noindent
\textbf{Problem Definition.}
The goal of the few-shot Generalist Anomaly Detection (GAD) task is to train a unified model that, given only a few normal images as visual prompts, can directly detect and localize anomalies from previously unseen classes.
Formally, let \(\mathcal{C}_{\text{seen}}\) denote the seen classes used for training and \(\mathcal{C}_{\text{unseen}}\) the unseen test classes, with \(\mathcal{C}_{\text{seen}} \cap \mathcal{C}_{\text{unseen}} = \emptyset\). The model is optimized on an auxiliary training set \(\mathcal{I}_{\text{train}}\) from \(\mathcal{C}_{\text{seen}}\), which contains both normal and anomalous images with image-level label \(\textbf{y}^{gt} \in \{0, 1\}\) and pixel-level ground-truth masks \(\textbf{M}^{gt} \in \{0,1\}^{H \times W}\), where \(H\) and \(W\) are the height and width of the image, respectively.
At inference, for any query image \(I_q \in \mathbb{R}^{H \times W \times 3}\)
from \(\mathcal{I}_{\text{test}}\), the model receives a few-shot visual prompt set
\(\mathcal{P}_v = \{I_p^{(1)}, \dots, I_p^{(K)}\}\) of \(K\) normal images of the
same unseen class. The model produces an image-level anomaly score \(\mathbf{s}\)
and a pixel-level anomaly score map \(\mathbf{M} \in \mathbb{R}^{H \times W}\).

\subsection{Rethinking CLIP for Few-shot GAD}
\label{sec:3.1}

\subsubsection{Dual-Branch Paradigm}

CLIP-based few-shot GAD methods typically follow a dual-branch paradigm.
Given a query image \(I_q\), the pre-trained CLIP visual encoder \(\mathcal{E}_v\)
augmented with V-V attention~\cite{li2023clip, zhou2023anomalyclip, li2024promptad} outputs the global \texttt{[CLS]} token
\(f_v^q \in \mathbb{R}^D\) and the patch-level feature map
\(F_v^q \in \mathbb{R}^{N \times D}\).
We use \(f \in \mathbb{R}^D\) as a generic query feature interchangeably
representing either the global token or a local patch token.
Throughout, \(\langle \cdot, \cdot \rangle\) denotes cosine similarity.
 
The \textbf{text branch} encodes normal and anomalous text prompts
\(\mathcal{P}_t^n\) and \(\mathcal{P}_t^a\) via the text encoder \(\mathcal{E}_t\),
averages them to obtain semantic anchors \(F_t^n\) and \(F_t^a\), and computes the
text-driven anomaly score via softmax:
\begin{equation}\label{eq:text_score}
\mathbf{S}_{\text{text}}
= \frac{\exp(\langle f, F_t^a \rangle)}
       {\exp(\langle f, F_t^n \rangle) + \exp(\langle f, F_t^a \rangle)}.
\end{equation}
The \textbf{visual branch} constructs a normal feature bank \(\mathcal{B}\) from
the few-shot visual prompt set \(\mathcal{P}_v\), retrieves the nearest reference
feature \(f_{\text{ref}}\) for each query patch, and computes:
\begin{equation}\label{eq:vis_score}
\mathbf{S}_{\text{vis}} = 1 - \langle f, f_{\text{ref}} \rangle.
\end{equation}

\subsubsection{Residual Representation in CLIP Feature Space}

A core challenge in few-shot GAD lies in decoupling anomaly-state semantics from
category-specific information. The original absolute feature space is dominated by
object identity, making direct anomaly measurement prone to cross-domain failure.
A relative representation \textit{residual} is therefore desired to cancel
out category-specific baselines while retaining state shifts.
 
Existing dual-branch paradigms already implicitly exploit residuals (See detailed Derivations in \textit{Appendix~\ref{app:derivation}}). Denoting
\(\ell_2\)-normalized features by \(\hat{\cdot}\), 
it can be shown
that the text-branch score is equivalently:
\begin{equation} \label{eq:text_score2residual}
\mathbf{S}_{\text{text}}
= \sigma\!\bigl(\|\mathbf{R}_t\|_2\,\langle f^q, \hat{\mathbf{R}}_t \rangle\bigr),
\quad
\mathbf{R}_t = \hat{F}_t^a - \hat{F}_t^n,
\end{equation}
where \(\sigma(\cdot)\) is the sigmoid function and \(\mathbf{R}_t\) is the
\textbf{text residual}. Since \(\|\mathbf{R}_t\|_2\) is a fixed constant, the
decision signal reduces to \(\langle f, \mathbf{R}_t \rangle\).
Similarly, the visual-branch score satisfies:
\begin{equation}
\mathbf{S}_{\text{vis}}
= \tfrac{1}{2}\|\mathbf{R}_v\|_2^2,
\quad
\mathbf{R}_v = \hat{f} - \hat{f}_{\text{ref}}.
\end{equation}
 
These equivalences reveal two structural limitations: 
(1) \textbf{Asymmetric modality alignment}: \(\mathbf{R}_t\) is forced to align with the raw visual feature \(f\), which varies significantly between foreground and background even for normal patches. This causes the alignment score \(\langle f, \mathbf{R}_t \rangle\) to drift with region type rather than anomaly severity, manifesting as the cross-granularity mismatch identified in ~\cref{sec:intro}.
\textbf{(2)~Loss of directional information}: the score
\(\|\mathbf{R}_v\|_2^2\) 
discards the high-dimensional directional information, which encodes the specific semantic nature of anomalies. This observation directly motivates our symmetric architecture that explicitly leverages residual features from both 
modalities.

\begin{figure}
    \centering
    \includegraphics[width=0.95\linewidth]{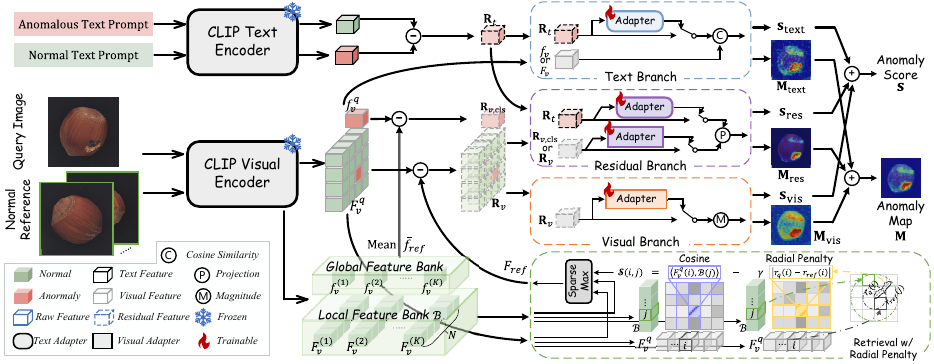}
    \caption{Overall architecture of Res\(^2\)CLIP, which consists of three branches in the residual domain. 
    }
    \label{fig:framework}
\end{figure}

\subsection{Res\(^2\)CLIP}

\subsubsection{Overall Architecture}

To fully leverage residual representations in CLIP feature space, we propose
Res\(^2\)CLIP, a three-branch framework that transfers CLIP's multimodal
alignment from the absolute feature space, dominated by class-specific information,
to a unified residual domain.
As illustrated in ~\cref{fig:framework}, we reconstruct the original text and
visual branches from a residual perspective and introduce a novel \textbf{residual
branch}. This branch enables direct and symmetric semantic alignment between visual
and textual modalities in the pure residual domain. It thereby resolves the asymmetry issue of aligning text residuals with raw visual features and preserves the directional information embedded in the residual features.
 
To support both training-free and fine-tuning settings, we introduce a unified
residual adaptation function \(\mathcal{A}(\cdot)\) that maps each raw residual
\(\mathbf{R}\) to an adapted residual \(\widetilde{\mathbf{R}} = \mathcal{A}(\mathbf{R})\).
In training-free mode, \(\mathcal{A}\) is the identity mapping. In fine-tuning mode,
it is instantiated as fully-connected layers for global features and multi-layer
convolutions for local features, with branch-specific adapters trained under
distinct objectives detailed below. Architecture details are
provided in \textit{Appendix~\ref{app:adapter}}.
 
The model aggregates the outputs of the three branches
at both image and pixel levels:
\begin{equation}
\mathbf{M} = \lambda^{\text{local}}_{\text{text}} \mathbf{M}_{\text{text}}
           + \lambda^{\text{local}}_{\text{vis}}  \mathbf{M}_{\text{vis}}
           + \lambda^{\text{local}}_{\text{res}}  \mathbf{M}_{\text{res}},
           \quad
\mathbf{s} = \lambda^{\text{global}}_{\text{text}} \mathbf{s}_{\text{text}}
           + \lambda^{\text{global}}_{\text{vis}}  \mathbf{s}_{\text{vis}}
           + \lambda^{\text{global}}_{\text{res}}  \mathbf{s}_{\text{res}},
\end{equation}
where \(\mathbf{M}_{\text{branch}} \in \mathbb{R}^{H \times W}\) and
\(\mathbf{s}_{\text{branch}} \in \mathbb{R}\)
(\(\text{branch} \in \{\text{text}, \text{vis}, \text{res}\}\)) denote the
pixel-level anomaly map and image-level anomaly score of each branch, and
\(\lambda^{\text{local}}_{\text{branch}}\), \(\lambda^{\text{global}}_{\text{branch}}\)
are the corresponding fusion weights at the pixel and image levels.

\subsubsection{Text Branch}

\noindent
\textbf{Text Residual Construction.}
We explicitly extract the text residual \(\mathbf{R}_t = \hat{F}_t^a - \hat{F}_t^n\),
which strips away class-shared attributes and preserves a pure anomaly
direction. The raw residual is then mapped to an adapted residual
\(\widetilde{\mathbf{R}}_t = \mathcal{A}_t(\mathbf{R}_t)\) via a lightweight
adapter \(\mathcal{A}_t\), refining the anomaly direction so that visual features
within seen training categories better align with it. 
Operating directly on the residual simplifies optimization compared with prior prompt learning methods that maintain separate adapters for normal and anomalous prompts.

\noindent
\textbf{Anomaly Scoring.}
Given a set of multi-layer local feature maps \(\{F_{v,l}^q\}_{l=1}^L\) extracted
from \(L\) different network layers, the cosine similarities with
\(\widetilde{\mathbf{R}}_t\) are averaged across layers and upsampled to obtain
the pixel-level anomaly map \(\mathbf{M}_{\text{text}} \in \mathbb{R}^{H \times W}\):
\begin{equation}
\mathbf{M}_{\text{text}}
= \operatorname{Upsample}\!
\left(
    \frac{1}{L} \sum_{l=1}^{L}
    \langle F_{v,l}^q,\, \widetilde{\mathbf{R}}_t \rangle
  \right),
\end{equation}
where \(\operatorname{Upsample}(\cdot)\) denotes bilinear interpolation. The
image-level score combines the \texttt{[CLS]} similarity with the mean of the top
\(1\%\) values in \(\mathbf{M}_{\text{text}}\):
\begin{equation}
\mathbf{s}_{\text{text}}
= \tfrac{1}{2}\langle f_v^q, \widetilde{\mathbf{R}}_t \rangle
+ \tfrac{1}{2}\operatorname{Mean}\!\bigl(\operatorname{Top}_{1\%}(\mathbf{M}_{\text{text}})\bigr),
\end{equation}
where \(\operatorname{Mean}(\cdot)\) denotes the mean operation and
\(\operatorname{Top}_{1\%}(\cdot)\) selects the highest \(1\%\) values.

\subsubsection{Visual Branch} \label{sec:visual_branch}

\noindent
\textbf{Visual Residual Construction with Radial Penalty.}
For the \(K\) normal reference images in \(\mathcal{P}_v\), we extract multi-layer
local features and spatially concatenate them across all images to construct the
\(l\)-th layer memory bank \(\mathcal{B}_l = \bigcup_{k=1}^K F_{v,l}^{(k)}\).
For a query feature \(F_{v,l}^q(i)\) at spatial position \(i\), naive
nearest-neighbor retrieval is prone to a critical risk: an anomalous patch may
match a structurally misaligned but semantically similar normal region,
contaminating the resulting visual residual. We mitigate this with a
\textbf{radial penalty} based on the structural prior that corresponding parts of
the same object lie at similar radial distances from the image center:
\begin{equation}
\mathcal{S}_l(i, j)
= \langle F_{v,l}^q(i),\, \mathcal{B}_l(j) \rangle
- \gamma  |r_q(i) - r_{\text{ref}}(j)|,
\end{equation}
where \(\mathcal{B}_l(j)\) is the \(j\)-th reference feature, \(r_q(i)\) and
\(r_{\text{ref}}(j)\) are the radial distances of the corresponding patch centers
from the image center, and \(\gamma\) controls the penalty strength. 
 
We apply Sparsemax~\cite{martins2016softmax} to \(\mathcal{S}_l\) to obtain adaptive sparse matching weights~\cite{qu2025dictas}, perform weighted aggregation over \(\mathcal{B}_l\) to yield a spatially
aligned reference feature \(F_{\text{ref},l}\), and extract the raw visual
residual \(\mathbf{R}_{v,l} = \hat{F}_{v,l}^q - \hat{F}_{\text{ref},l}\). The raw residual is then mapped to an adapted residual \(\widetilde{\mathbf{R}}_{v,l} = \mathcal{A}_{v,l}(\mathbf{R}_{v,l})\) via a layer-specific adapter that acts as a denoising filter, suppressing residual noise in normal regions while preserving the magnitude of genuine anomalies.

\noindent
\textbf{Anomaly Scoring.}
The pixel-level anomaly map \(\mathbf{M}_{\text{vis}} \in \mathbb{R}^{H \times W}\) and the image-level anomaly score \(s_{\text{vis}}\) are directly computed from the magnitudes of the adapted residuals across all layers:
\begin{equation}
\textbf{M}_{\text{vis}} = \text{Upsample}\left( \frac{1}{L} \sum_{l=1}^L \|\widetilde{\mathbf{R}}_{v,l}\|_2^2 \right),
\quad
\textbf{s}_{\text{vis}} = \text{Mean}(\text{Top}_{1\%}(M_{\text{vis}})).
\end{equation}

\subsubsection{Residual Branch}
\label{sec:residual_branch}
\noindent
\textbf{Residual-to-Residual Alignment \(\mathbf{R}_v \leftrightarrow \mathbf{R}_t\).}
We propose to align the visual residual and the text residual directly within the
residual domain. The key insight is the following equivalence:
\begin{equation}\label{eq:proj}
\langle f^q, \mathbf{R}_t \rangle - \langle f_{\text{ref}}, \mathbf{R}_t \rangle
= \mathbf{R}_v \cdot \hat{\mathbf{R}}_t,
\end{equation}
where \(\cdot\) denotes the dot product along the feature channel dimension.
Calibrating the text-branch score with a normal reference is therefore mathematically equivalent to projecting the visual residual onto the normalized text residual direction, a fully symmetric operation in the residual domain. This formulation simultaneously addresses both limitations identified in ~\cref{sec:3.1}: the region-type baseline drift is eliminated by replacing the absolute feature \(f^q\) with the differential \(\mathbf{R}_v\), and the high-dimensional directional information of \(\mathbf{R}_v\), previously discarded by the magnitude-only score \(|\mathbf{R}_v|_2^2\), is now fully exploited through the dot product. 
We use the dot product rather than cosine similarity to preserve the magnitude of
\(\mathbf{R}_v\) as an anomaly severity signal. See \textit{Appendix~\ref{app:symmetric_alignment}} and \textit{~\ref{app:projection_vs_cosine}} for the derivation and numerical justification.

\noindent
\textbf{Residual Construction.}
We reuse \(\mathbf{R}_t\) from the text branch and
\(\{\mathbf{R}_{v,l}\}_{l=1}^L\) from the visual branch. To supplement the
image-level understanding, we average the global features of the normal reference
images to build the normal baseline
\(\bar{f}_{\text{ref}} = \tfrac{1}{K}\sum_{k=1}^K f_v^{(k)}\), and extract the
global visual residual \(\mathbf{R}_{v,\text{cls}} = f_v^q - \bar{f}_{\text{ref}}\).
While these raw residuals provide a solid theoretical foundation, CLIP's pre-trained features are not specifically aligned with anomaly detection in specialized domains such as industrial inspection. We therefore introduce three types of independent adapters \(\mathcal{A}_t^{\text{res}}\), \(\{\mathcal{A}_{v,l}^{\text{res}}\}_{l=1}^L\), and \(\mathcal{A}_{v,\text{cls}}^{\text{res}}\) to map the raw residuals into an adapted residual space, yielding \(\widetilde{\mathbf{R}}_t^{\text{res}}\), \(\{\widetilde{\mathbf{R}}_{v,l}^{\text{res}}\}_{l=1}^L\), and
\(\widetilde{\mathbf{R}}_{v,\text{cls}}^{\text{res}}\) that better fit the distribution requirements of anomaly detection.

\noindent
\textbf{Anomaly Scoring via Projection.}
The pixel-level anomaly map \(\mathbf{M}_{\text{res}} \in \mathbb{R}^{H \times W}\) aggregates the projection scores between the local
adapted visual residuals and the normalized adapted text residual
\(\hat{\widetilde{\mathbf{R}}}_t^{\text{res}}\) across layers:
\begin{equation}
\mathbf{M}_{\text{res}}
= \operatorname{Upsample}\!\left(
    \frac{1}{L} \sum_{l=1}^{L}
    \widetilde{\mathbf{R}}_{v,l}^{\text{res}}
    \cdot
    \hat{\widetilde{\mathbf{R}}}_t^{\text{res}}
  \right).
\end{equation}
The image-level score fuses the global residual projection with the mean of the
top \(1\%\) values in \(\mathbf{M}_{\text{res}}\):
\begin{equation}
\mathbf{s}_{\text{res}}
= \tfrac{1}{2}
  \widetilde{\mathbf{R}}_{v,\text{cls}}^{\text{res}}
  \cdot
  \hat{\widetilde{\mathbf{R}}}_t^{\text{res}}
+ \tfrac{1}{2}
  \operatorname{Mean}\!\bigl(\operatorname{Top}_{1\%}(\mathbf{M}_{\text{res}})\bigr).
\end{equation}

\begin{figure}
    \centering
    \includegraphics[width=\linewidth]{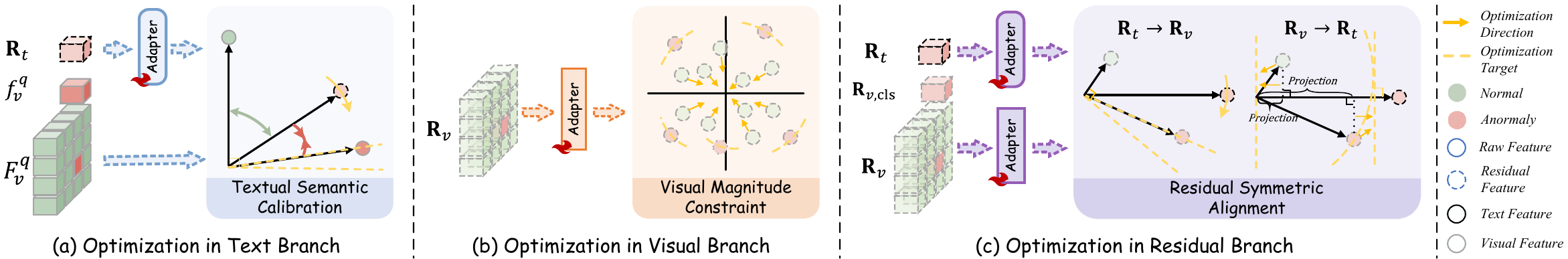}
    \caption{Optimization objectives for the three branches. (a) Text branch: the adapted text residual is pushed toward anomalous visual features and away from normal ones. (b) Visual branch: adapted residual magnitudes are suppressed to zero for normal patches and preserved for anomalous patches. (c) Residual branch: two adapters are alternately optimized via projection. Updating the text adapter (left) aligns the text residual with anomalous visual residuals; updating the visual adapter (right) drives anomalous projections toward the raw residual magnitude and normal projections toward zero.
    }
    \label{fig:optimization}
\end{figure}

\subsection{Optimization in Residual Space} \label{sec:optimization}

All learnable adapters are optimized within the pure residual domain, as illustrated in ~\cref{fig:optimization}. Unadapted
residuals and raw CLIP features are kept frozen throughout training.

\noindent
\textbf{Textual Semantic Calibration in Text Branch.}
To optimize \(\mathcal{A}_t\), we constrain
\(\langle f, \widetilde{\mathbf{R}}_t \rangle\) via a case-specific penalties losses: for normal
samples the similarity is pushed below 0, and for anomalous samples toward 1. A
regularization term prevents \(\widetilde{\mathbf{R}}_t\) from drifting from the
original CLIP semantic manifold. The full  formulations
\(\mathcal{L}_{\text{text}}^{\text{global}}\) and
\(\mathcal{L}_{\text{text}}^{\text{local}}\) are given in
\textit{Appendix~\ref{app:loss_text}}:
\begin{equation}
\mathcal{L}_{\text{text}}
= \mathcal{L}_{\text{text}}^{\text{global}}
+ \mathcal{L}_{\text{text}}^{\text{local}}
+ \bigl(1 - \langle \widetilde{\mathbf{R}}_t,\, \mathbf{R}_t \rangle\bigr).
\end{equation}

\noindent
\textbf{Visual Magnitude Constraint in Visual Branch.}
For \(\{\mathcal{A}_{v,l}\}_{l=1}^L\), the adapted residual magnitude is treated
as the degree of feature offset. Normal-pixel magnitudes are pushed toward 0 to
suppress retrieval noise, while anomalous-pixel magnitudes are constrained to
approximate those of the raw physical residuals \(\mathbf{R}_{v,l}\), preserving
the original anomaly signal:
\begin{equation}
\mathcal{L}_{\text{vis}}
= \sum_{l=1}^{L} \!\left(
    \frac{1}{N_0} \!\sum_{\mathbf{M}^{gt}=0}\! \|\widetilde{\mathbf{R}}_{v,l}\|_2
  + \frac{1}{N_1} \!\sum_{\mathbf{M}^{gt}=1}\!
    \bigl|\|\widetilde{\mathbf{R}}_{v,l}\|_2 - \|\mathbf{R}_{v,l}\|_2\bigr|
  \right),
\end{equation}
where \(N_0\) and \(N_1\) are the numbers of normal and anomalous pixels.

\noindent\textbf{Residual Symmetric Alignment in Residual Branch.}
For the residual branch, we adopt an alternating optimization strategy to prevent
feature collapse during joint training of \(\mathcal{A}_t^{\text{res}}\) and
\(\mathcal{A}_v^{\text{res}}\). In each iteration, one modality is frozen via the
stop-gradient operator \(\operatorname{sg}(\cdot)\) while the other is updated.
The per-sample alignment target uses the raw visual residual magnitude as a lower
bound for anomalous samples, since perfect alignment yields a projection equal to
that magnitude. Defining hinge-style base functions
\(\ell_{\text{res}}^{\text{global}}(P_{\text{cls}})\) and
\(\ell_{\text{res}}^{\text{local}}(P_l)\) over generic projection terms, and
regularization terms \(\mathcal{L}_{\text{reg}}^t\), \(\mathcal{L}_{\text{reg}}^v\)
that prevent features from deviating from the original manifolds (full forms in
\textit{Appendix~\ref{app:loss_res}}), the alternating objective is:
\begin{equation}
\mathcal{L}_{\text{res}} =
\begin{cases}
  \ell_{\text{res}}^{\text{global}}\!\bigl(
    \operatorname{sg}(\widetilde{\mathbf{R}}_{v,\text{cls}}^{\text{res}})
    \cdot \hat{\widetilde{\mathbf{R}}}_t^{\text{res}}
  \bigr)
  + \sum_{l=1}^L 
    \ell_{\text{res}}^{\text{local}}\!\bigl(
      \operatorname{sg}(\widetilde{\mathbf{R}}_{v,l}^{\text{res}})
      \cdot \hat{\widetilde{\mathbf{R}}}_t^{\text{res}}
    \bigr)
  + \mathcal{L}_{\text{reg}}^{t},
  & \text{updating } \mathcal{A}_t^{\text{res}} \\[8pt]
  \ell_{\text{res}}^{\text{global}}\!\bigl(
    \widetilde{\mathbf{R}}_{v,\text{cls}}^{\text{res}}
    \cdot \operatorname{sg}(\hat{\widetilde{\mathbf{R}}}_t^{\text{res}})
  \bigr)
  + \sum_{l=1}^{L}
    \ell_{\text{res}}^{\text{local}}\!\bigl(
      \widetilde{\mathbf{R}}_{v,l}^{\text{res}}
      \cdot \operatorname{sg}(\hat{\widetilde{\mathbf{R}}}_t^{\text{res}})
    \bigr)
  + \mathcal{L}_{\text{reg}}^{v},
  & \text{updating } \mathcal{A}_v^{\text{res}}
\end{cases}
\end{equation}

The overall training objective is:
\begin{equation}
\mathcal{L}_{\text{total}}
= \mathcal{L}_{\text{text}}
+ \mathcal{L}_{\text{vis}}
+ \mathcal{L}_{\text{res}}.
\end{equation}

\begin{table*}
  \caption{Comprehensive quantitative comparison on five anomaly detection datasets under 1, 2, and 4-shot setting (I-AUC / P-AP / PRO), with the best results in \textbf{bold} and the second-best \underline{underlined}. 
  }
  \label{tab:main_results}
  \centering
  \resizebox{0.9\linewidth}{!}{
  \begin{tabular}{c|c|cc|ccccc}
    \toprule
     \multirow{2}{*}{Shots}  & \multirow{2}{*}{Dataset} & \multicolumn{2}{c|}{\textit{Training-free}} & \multicolumn{5}{c}{\textit{Fine-tune}} \\
     \cmidrule(r){3-9}
     &   & WinCLIP+~\cite{jeong2023winclip}  & Res\(^2\)CLIP\(^*\) (Ours) & APRIL-GAN~\cite{chen2023zero}  & AnomalyCLIP+~\cite{zhou2023anomalyclip} & ReMP-AD~\cite{ma2025remp} & AdaptCLIP~\cite{gao2026adaptclip}  & Res\(^2\)CLIP\(^\dagger\) (Ours)\\
    \midrule
   \multirow{5}{*}{1}  & MVTecAD 
   & 93.1 / 38.5 / 84.1 & \textbf{96.1} / 50.8 / 91.1 & 92.2 / 52.5 / 90.8 & 93.4 / 46.2 / 89.5 & 95.2 / \underline{57.3} / \underline{91.6} & 95.0 / 53.9 / 89.7 & \underline{96.0} / \textbf{57.9} / \textbf{91.7} \\
                       & VisA & 82.9 / 15.5 / 80.2 & 88.2 / 28.2 / 89.6 & \underline{91.6} / 31.0 / 90.1 & 80.0 / 28.2 / 89.6 & \textbf{92.1} / 35.3 / \underline{91.8} & 91.2 / \textbf{40.0} / 91.2 & 89.5 / \underline{39.2} / \textbf{92.0} \\
                      & BTAD & 85.0 / 42.0 / 66.7 & \underline{95.5} / \underline{63.0} / \underline{82.8} & 91.7 / 50.1 / 78.6 & 92.7 / 58.4 / 79.4 & 94.9 / 53.9 / 82.3 & 93.1 / 61.4 / 77.7 & \textbf{95.9} / \textbf{64.2} / \textbf{83.0} \\
                       & MPDD & 69.1 / 30.7 / 88.4 & \textbf{86.6} / 36.9 / \underline{94.0} & 82.6 / 35.6 / 91.2 & 79.7 / 36.8 / 93.5 & 82.8 / \textbf{38.9} / 93.7 & \underline{84.1} / 34.7 / 93.7 & 83.6 / \underline{37.3} / \textbf{94.2} \\
                       & DTD & 98.1 / 48.8 / 90.3 & 95.5 / 69.5 / 90.3 & \underline{98.4} / 76.5 / 92.0 & 95.3 / 70.6 / 91.5 & \textbf{99.3} / \textbf{79.2} / 92.5 & 98.0 / 76.9 / \underline{93.1} & 97.2 / \underline{77.5} / \textbf{94.6} \\
                       \cmidrule(r){2-9}
                       & \cellcolor{gray!10}\textbf{Average} & \cellcolor{gray!10}85.6 / 35.1 / 81.9 & \cellcolor{gray!10}\underline{92.4} / 49.7 / 89.6 & \cellcolor{gray!10}91.3 / 49.1 / 88.5 & \cellcolor{gray!10}88.2 / 48.0 / 88.7 & \cellcolor{gray!10}\textbf{92.9} / 52.9 / \underline{90.4} & \cellcolor{gray!10}92.3 / \underline{53.4} / 89.1 & \cellcolor{gray!10}\underline{92.4} / \textbf{55.2} / \textbf{91.1} \\
                                    \midrule
   \multirow{5}{*}{2} & MVTecAD & 94.7 / 40.0 / 84.6 & \underline{96.3} / 53.3 / \underline{92.0} & 92.4 / 53.5 / 91.3 & 93.7 / 47.7 / 90.2 & 95.2 / \underline{57.2} / 91.9 & 95.7 / 55.5 / 90.3 & \textbf{96.9} / \textbf{60.6} / \textbf{92.7} \\
                       & VisA & 83.5 / 16.5 / 80.8 & 89.3 / 30.3 / 90.3 & 92.3 / 31.6 / 90.1 & 82.2 / 28.8 / 89.4 & \underline{92.6} / 35.1 / \underline{91.9} & \textbf{92.7} / \textbf{41.1} / 91.6 & 90.4 / \underline{40.6} / \textbf{92.7} \\
                      & BTAD & 85.7 / 43.3 / 66.7 & \underline{95.7} / \underline{64.5} / \underline{83.1} & 91.9 / 50.6 / 78.3 & 92.5 / 59.1 / 79.0 & 95.2 / 54.4 / 82.3 & 92.9 / 61.7 / 77.9 & \textbf{96.2} / \textbf{65.7} / \textbf{83.5} \\
                       & MPDD & 70.1 / 31.9 / 89.5 & \textbf{87.6} / 38.3 / \underline{94.5} & 84.6 / 36.9 / 91.3 & 78.1 / 37.6 / 94.1 & \underline{87.0} / \textbf{41.4} / 93.8 & 85.4 / 35.7 / 94.3 & 85.8 / \underline{39.0} / \textbf{94.7} \\
                       & DTD & 98.2 / 49.3 / 90.4 & 95.9 / 71.2 / 90.8 & \underline{98.4} / 76.9 / 92.2 & 95.0 / 71.2 / 91.7 & \textbf{99.4} / \textbf{80.7} / 93.1 & 98.2 / 78.0 / \underline{93.2} & 97.4 / \underline{78.4} / \textbf{94.7} \\
                       \cmidrule(r){2-9}
                       & \cellcolor{gray!10}\textbf{Average} & \cellcolor{gray!10}86.4 / 36.2 / 82.4 & \cellcolor{gray!10}93.0 / 51.5 / 90.1 & \cellcolor{gray!10}91.9 / 49.9 / 88.6 & \cellcolor{gray!10}88.3 / 48.9 / 88.9 & \cellcolor{gray!10}\textbf{93.9} / 53.8 / \underline{90.6} & \cellcolor{gray!10}93.0 / \underline{54.4} / 89.5 & \cellcolor{gray!10}\underline{93.3} / \textbf{56.9} / \textbf{91.7} \\
                                    \midrule
   \multirow{5}{*}{4} & MVTecAD & 95.1 / 42.0 / 85.5 & \underline{97.5} / 55.3 / \underline{92.8} & 92.7 / 54.8 / 91.9 & 93.8 / 48.6 / 90.7 & 95.8 / \underline{58.0} / 92.3 & 96.8 / 57.8 / 91.1 & \textbf{97.9} / \textbf{62.5} / \textbf{93.5} \\
                       & VisA & 84.4 / 17.9 / 81.3 & 90.1 / 31.4 / 90.8 & 92.8 / 31.8 / 90.1 & 81.7 / 29.3 / 89.7 & \textbf{93.3} / 34.8 / 91.9 & \textbf{93.3} / \textbf{42.0} / \underline{92.0} & 91.1 / \underline{41.0} / \textbf{93.3} \\
                      & BTAD & 86.4 / 45.0 / 67.0 & \underline{96.4} / \underline{65.6} / \underline{83.2} & 91.8 / 51.1 / 78.5 & 92.4 / 59.3 / 79.1 & 94.4 / 55.6 / 82.0 & 93.1 / 62.8 / 78.3 & \textbf{96.8} / \textbf{67.1} / \textbf{83.5} \\
                       & MPDD & 71.8 / 32.2 / 88.5 & \underline{89.0} / 40.0 / \underline{95.0} & 85.9 / 37.3 / 91.1 & 78.4 / 38.6 / 94.2 & \textbf{89.6} / \textbf{43.1} / 93.6 & 86.3 / 39.5 / 94.9 & 87.3 / \underline{42.6} / \textbf{95.4} \\
                       & DTD & 98.3 / 50.0 / 90.7 & 96.9 / 73.2 / 92.2 & \underline{98.6} / 77.3 / 92.5 & 95.3 / 71.8 / 92.2 & \textbf{99.4} / \textbf{81.2} / 93.5 & 98.2 / 78.4 / \underline{93.6} & 97.7 / \underline{79.2} / \textbf{95.0} \\
                       \cmidrule(r){2-9}
                       & \cellcolor{gray!10}\textbf{Average} & \cellcolor{gray!10}87.2 / 37.4 / 82.6 & \cellcolor{gray!10}94.0 / 53.1 / \underline{90.8} & \cellcolor{gray!10}92.4 / 50.5 / 88.8 & \cellcolor{gray!10}88.3 / 49.5 / 89.2 & \cellcolor{gray!10}\textbf{94.5} / 54.5 / 90.7 & \cellcolor{gray!10}93.5 / \underline{56.1} / 90.0 & \cellcolor{gray!10}\underline{94.2} / \textbf{58.5} / \textbf{92.1} \\
                                    \midrule
    \bottomrule
  \end{tabular}
  }
\end{table*}

\section{Experiments}
\label{sec:ex}

\subsection{Experimental Setup}  \label{sec:exsetup}

\noindent
\textbf{Datasets.}
Our evaluation covers five anomaly detection datasets: MVTecAD~\cite{bergmann2019mvtec},
VisA~\cite{zou2022spot}, BTAD~\cite{mishra2021vt}, MPDD~\cite{jezek2021deep}, and DTD-Synthetic~\cite{aota2023zero} (abbreviated as DTD).
Following the few-shot generalist anomaly detection protocol, we use the
test set of MVTecAD as the auxiliary training set for all target datasets,
and switch to the test set of VisA when MVTecAD itself is the target. This
ensures that no images from the target domain are seen during training.

\noindent
\textbf{Evaluation Metrics.}
For each setting we evaluate both anomaly classification and anomaly
localization. Image-level performance is reported with the area under the
ROC curve (I-AUC). Pixel-level performance is reported with the average precision (P-AP), and the per-region
overlap (PRO). 

\noindent
\textbf{Implementation Details.}
We adopt the CLIP model ViT-L/14@336px pretrained by OpenAI~\cite{radford2021learning} as the backbone, with
V-V attention~\cite{li2023clip} applied to the last 20 transformer layers of the visual
encoder. The CLIP backbone is kept frozen throughout training. The input images are all resized to \(518 \times 518\). We extract
patch features from layers \(\{6, 12, 18, 24\}\) (\(L = 4\)). The radial penalty weight is set to
\(\gamma = 0.01\). 
All experiments are run on a single NVIDIA RTX 3090.

\subsection{Main Results}

\begin{figure}
    \centering
    \includegraphics[width=0.9\linewidth]{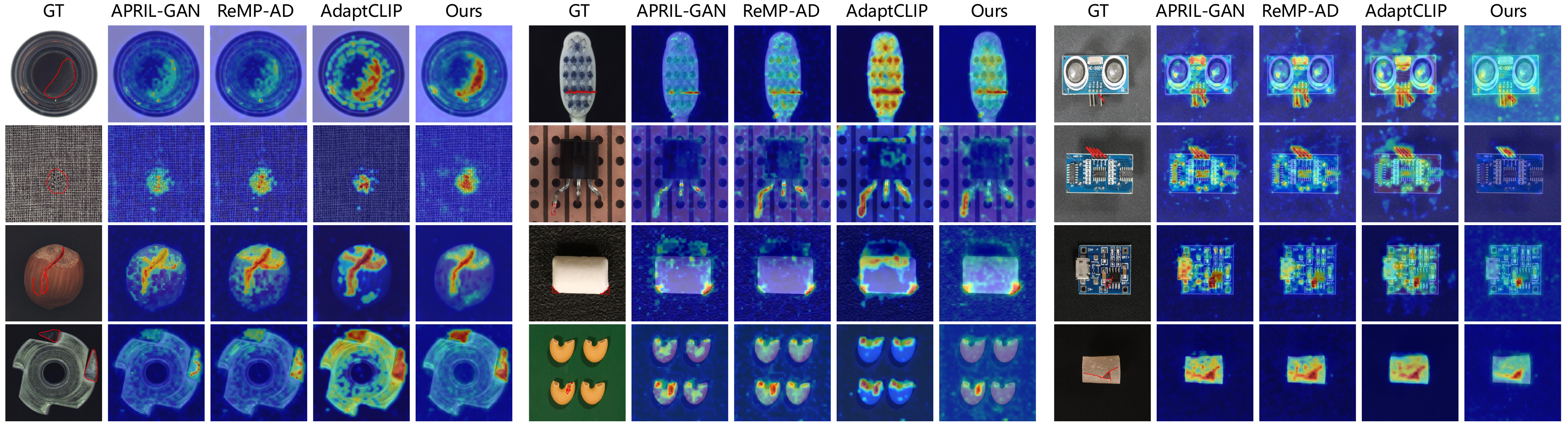}
    \caption{Anomaly map comparison with state‑of‑the‑art methods under the 1‑shot setting.
    }
    \label{fig:map_results}
\end{figure}

We evaluate two settings of our method, \textbf{Res\(^2\)CLIP\(^*\)} and
\textbf{Res\(^2\)CLIP\(^\dagger\)}, corresponding to the training-free mode
where the adapter \(\mathcal{A}(\cdot)\) is the identity mapping, and the
fine-tune mode where the adapters are optimized using the auxiliary training
set, respectively.

\noindent
\textbf{Quantitative Comparison.}
\cref{tab:main_results} compares Res\(^2\)CLIP with state-of-the-art methods
across five datasets under 1-, 2-, and 4-shot settings. On the localization
metrics most relevant to industrial inspection,
Res\(^2\)CLIP\(^\dagger\) achieves the highest average P-AP and PRO across
all three shot settings, outperforming the next-best method by
+1.8\%/+2.0\%/+2.4\% on P-AP and +0.7\%/+1.1\%/+1.3\% on PRO for 1/2/4
shots respectively. The gains are particularly pronounced on BTAD, where
P-AP exceeds the second-best by over 2.8\% under every shot count,
reflecting the benefit of residual-domain alignment on datasets with
complex industrial textures. On image-level I-AUC, Res\(^2\)CLIP performs
competitively and achieves the best results on MVTecAD and BTAD across all
settings. Notably, our training-free Res\(^2\)CLIP\(^*\) already surpasses
several fine-tuned methods on localization, confirming that the residual-to-residual alignment inherently provides a strong inductive bias for anomaly localization, independent of subsequent adaptation.
 
\noindent
\textbf{Qualitative Comparison.}
\cref{fig:map_results} shows representative anomaly
maps on MVTecAD and VisA under the 1-shot setting.
Res\(^2\)CLIP consistently produces tighter localization around the ground-truth defect regions, demonstrating superior localization precision.

\subsection{Ablation Studies}
\label{sec:ablation}
All ablation experiments are conducted under the 1-shot setting, and for each dataset we keep the same normal reference image across all comparisons to ensure fairness.

\begin{figure}[t]
    \centering
    
    \begin{minipage}[t]{0.63\linewidth}
    \vspace{0pt}
        \centering
        \captionof{table}{Comprehensive ablation study of different components. }
        \label{tab:comprehensive_ablation}
        \resizebox{\columnwidth}{!}{
            \begin{tabular}{l|l|ccccc|ccccc}
\toprule
\multicolumn{2}{l|}{Mode} & \multicolumn{5}{c|}{Training-free} & \multicolumn{5}{c}{Fine-tune} \\
\midrule
\multicolumn{2}{l|}{Text Branch (\textbf{T})} & \checkmark & & & \checkmark & \checkmark & \checkmark&  &  &\checkmark & \checkmark \\
\multicolumn{2}{l|}{Visual Branch (\textbf{V})} & & \checkmark & & \checkmark &\checkmark &  & \checkmark & & \checkmark & \checkmark \\
\multicolumn{2}{l|}{Residual Branch (\textbf{R})} & & & \checkmark & & \checkmark &  & & \checkmark &  & \checkmark \\
\multicolumn{2}{l|}{Adapter} & & & & & & \checkmark& \checkmark & \checkmark & \checkmark & \checkmark \\
\midrule
\multirow{3}{*}{MVTec AD}
& I-AUC & 91.5 & 91.1 & 94.3 & 94.4 & \textbf{96.4} & 92.3 & 94.1 & 92.9 & 96.1 & 96.1 \\
& P-AP  & 30.9 & 53.7 & 33.3 & 53.7 & 50.3 & 32.4 & 56.4 & 50.4 & 56.1 & \textbf{57.4} \\
& PRO   & 83.2 & 88.2 & 84.7 & 88.9 & 90.8 & 83.8 & 90.9 & 87.9 & 91.2 & \textbf{91.4} \\
\midrule
\multirow{3}{*}{VisA}
& I-AUC & 82.8 & 86.8 & 84.4 & 88.7 & 88.5 & 82.9 & 88.6 & 87.0 & \textbf{89.8} & \textbf{89.8} \\
& P-AP  & 16.5 & 31.2 & 16.5 & 31.8 & 28.2 & 17.4 & 39.0 & 32.7 & 38.8 & \textbf{39.5} \\
& PRO   & 85.0 & 87.6 & 79.6 & 88.3 & 89.6 & 86.7 & 91.4 & 88.0 & 91.8 & \textbf{92.1} \\
\midrule
\multirow{3}{*}{BTAD}
& I-AUC & 91.9 & 93.7 & 92.9 & 94.7 & 95.6 & 91.9 & 94.9 & 94.1 & \textbf{95.7} & \textbf{95.7} \\
& P-AP  & 43.8 & 62.5 & 51.1 & 63.0 & 64.2 & 44.6 & 61.9 & 63.4 & 62.5 & \textbf{65.3} \\
& PRO   & 78.2 & 81.2 & 80.7 & 81.5 & 83.0 & 78.3 & 82.2 & 83.4 & 82.6 & \textbf{83.1} \\
\midrule
\rowcolor{gray!10} 
& I-AUC & 88.7 & 90.5 & 90.5 & 92.6 & 93.5 & 89.0 & 92.5 & 91.3 & 93.9 & \textbf{93.9} \\
\rowcolor{gray!10}
& P-AP  & 30.4 & 49.1 & 33.6 & 49.5 & 47.6 & 31.5 & 52.4 & 48.8 & 52.5 & \textbf{54.1} \\
\rowcolor{gray!10}
\multirow{-3}{*}{\textbf{Average}}
& PRO   & 82.1 & 85.7 & 81.7 & 86.2 & 87.8 & 82.9 & 88.2 & 86.4 & 88.5 & \textbf{88.9} \\
\bottomrule
\end{tabular}
        }
    \end{minipage}
    \hfill 
    \begin{minipage}[t]{0.35\linewidth}
    \vspace{0pt}
        \centering
        \includegraphics[width=\linewidth]{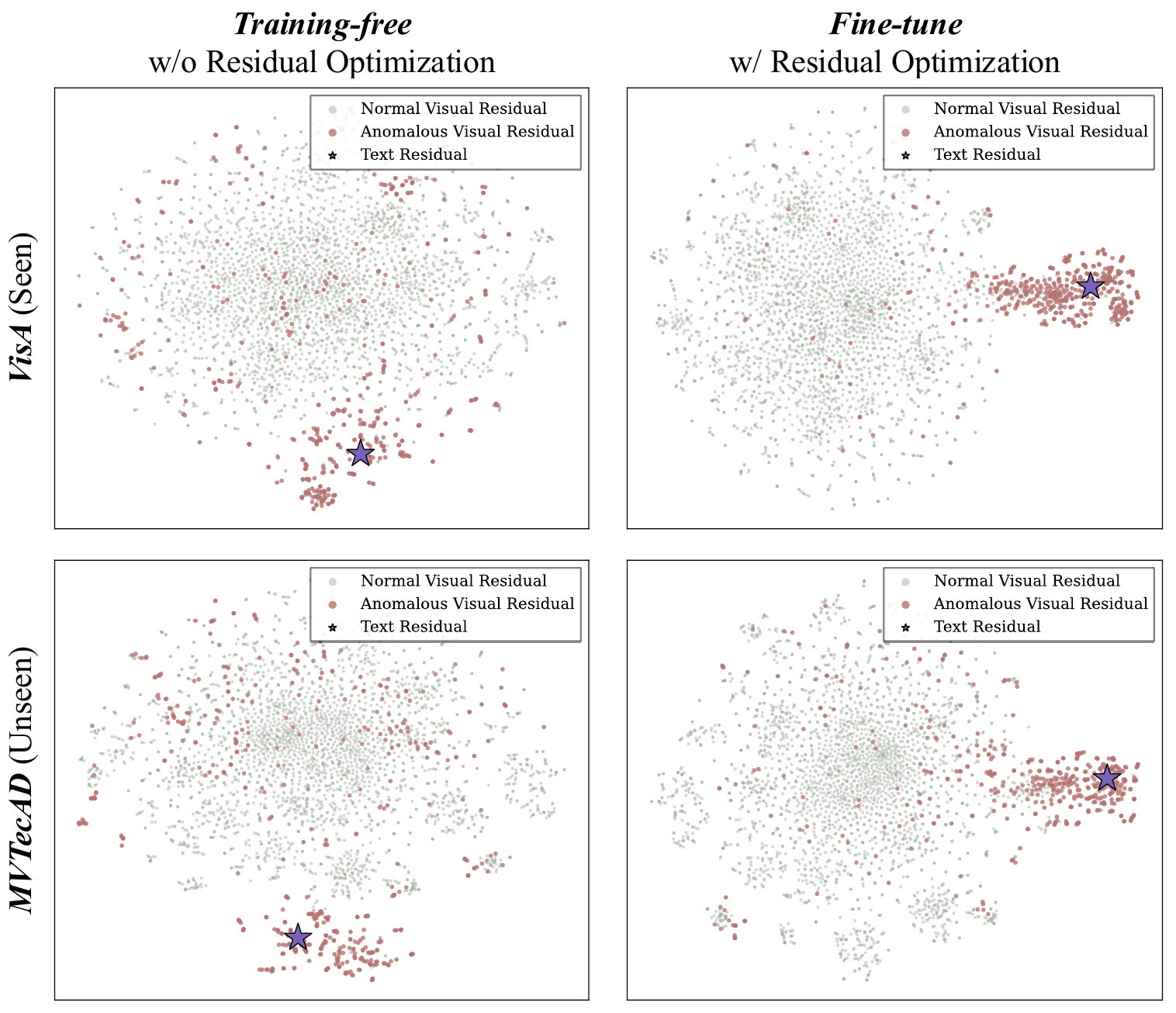} 
        \captionof{figure}{t-SNE visualization of the residual features of two modalities before and after optimization.}
        \label{fig:adapter_main}
    \end{minipage}
    
\end{figure}

\noindent
\textbf{Effectiveness of Architecture Design.} 
To validate the independent contributions of each branch, we test different branch combinations in \cref{tab:comprehensive_ablation}. In the training-free setting, the traditional text and visual dual-branch (\textbf{T}+\textbf{V}) achieves basic detection performance. Introducing the residual branch (\textbf{T}+\textbf{V}+R) improves the training-free baseline, reaching an average I-AUC of 93.5\% (+0.9\% over \textbf{T}+\textbf{V}) and an average PRO of 87.8\% (+1.6\%). 
Under the fine-tune setting, the complete \textbf{T}+\textbf{V}+\textbf{R} architecture again achieves the best average performance across all metrics.
These results indicate that the three-branch design is consistently beneficial across both modes.

\noindent
\textbf{Effectiveness of Residual Optimization.}
Adding adapters yields consistent gains across all branch configurations,
and the largest gains occur on the residual branch. Comparing the
training-free and fine-tune settings reveals an interesting phenomenon: the
raw visual residuals obtained by direct subtraction in the training-free
setting unavoidably carry spatial mismatch noise from imperfect retrieval,
which causes a temporary decrease on the false-positive-sensitive P-AP
metric (e.g., from 53.7\% to 50.3\% on MVTecAD when the unadapted residual
branch is added). However, constraining the optimization within the
residual domain effectively resolves this issue. After fine-tuning with
adapters (\textbf{T}+\textbf{V}+\textbf{R}+Adapter), the model reaches the
highest P-AP on every dataset, with the average P-AP increasing from the
training-free baseline of 47.6\% to 54.1\% (+6.5\%).

\cref{fig:adapter_main} visualizes the residual feature distributions before and after adaptation. Before optimization, the anomalous visual residuals are scattered and entangled with normal residuals. After adaptation, anomalous residuals form a tighter cluster around the text residual anchor, and the separation between normal and anomalous residuals becomes substantially clearer, illustrating the role of the residual adapter in suppressing retrieval noise and enhancing anomaly alignment.

\begin{table}[t]
    \centering
	\begin{minipage}[t]{0.42\linewidth}
\caption{Comparison of different residual alignment metrics in the residual branch in training-free setting.}
  \label{tab:metric_ablation}
  \centering
  \resizebox{0.78\columnwidth}{!}{
  \begin{tabular}{ll|cc}
\toprule
\multirow{2}{*}{Dataset} & \multirow{2}{*}{Metric} & Cos. & Proj.\\
& & $\langle \mathbf{R}_v, \mathbf{R}_t \rangle$ & $\mathbf{R}_v \cdot \hat{\mathbf{R}}_t$ \\
\midrule
\multirow{3}{*}{MVTec AD} 
& I-AUC & 91.7 & 94.3 \\
& P-AP  & 27.5 & 33.3 \\
& PRO   & 79.2 & 84.7 \\
\midrule
\multirow{3}{*}{VisA}     
& I-AUC & 80.6 & 84.4 \\
& P-AP  & 12.0 & 16.5 \\
& PRO   & 73.4 & 79.6 \\
\midrule
\multirow{3}{*}{BTAD}     
& I-AUC & 87.1 & 92.9 \\
& P-AP  & 36.6 & 51.1 \\
& PRO   & 75.1 & 80.7 \\
\bottomrule
\end{tabular}
  }
	    \end{minipage}
\hfill
        \begin{minipage}[t]{0.55\linewidth}
\caption{Ablation of the radial penalty on MVTec AD. TF and FT denote the training-free and fine-tune settings, respectively.}
\vspace{2mm}
  \label{tab:radial_mvtec}
  \centering
  \resizebox{0.85\columnwidth}{!}{
\begin{tabular}{cl|c | ccc}
\toprule
Mode & Branch & Radial Penalty & I-AUC & P-AP & PRO \\
\midrule
\multirow{4}{*}{TF} 
& \multirow{2}{*}{Visual}   & & 90.7 & 53.1 & 87.7 \\
&                           & \checkmark & 91.1 & 53.7 & 88.2 \\
\cmidrule(lr){2-6}
& \multirow{2}{*}{Residual} & & 94.0 & 32.9 & 84.4 \\
&                           & \checkmark & 94.3 & 33.3 & 84.7 \\
\midrule
\multirow{4}{*}{FT}     
& \multirow{2}{*}{Visual}   & & 93.4 & 55.7 & 90.5 \\
&                           & \checkmark & 94.1 & 56.4 & 90.9 \\
\cmidrule(lr){2-6}
& \multirow{2}{*}{Residual} & & 92.2 & 49.1 & 87.4 \\
&                           & \checkmark & 92.9 & 50.4 & 87.9 \\
\bottomrule
\end{tabular}
  }
	    \end{minipage}
\end{table}

\noindent
\textbf{Alignment Metric: Projection vs. Cosine Similarity.}
We compare vector projection with cosine similarity for residual alignment in \cref{tab:metric_ablation}. Across all three datasets, projection consistently outperforms cosine similarity on every metric. The
improvements are particularly evident in localization: projection raises
P-AP from 27.5\% to 33.3\% on MVTecAD, from 12.0\% to 16.5\% on VisA, and
from 36.6\% to 51.1\% on BTAD. This advantage stems from the fact that
cosine similarity normalizes small-magnitude background residuals and
amplifies their noise direction, whereas dot-product projection naturally
preserves magnitude, suppressing false positives in normal regions while
capturing the angular alignment of true anomalies.

\noindent
\textbf{Effectiveness of Radial Penalty.} 
We examine the impact of the radial penalty in
\cref{tab:radial_mvtec}. Across both training-free and fine-tune settings,
adding the radial penalty consistently brings steady improvements
on the visual and residual branches. The gains are most meaningful on the
residual branch under fine-tuning, where P-AP increases from 49.1\% to
50.4\%. This indicates that the radial penalty regularizes the retrieval
process and yields cleaner residuals for downstream pixel-level alignment.
 
\section{Conclusions} \label{conclusion}

We presented Res\(^2\)CLIP, a residual-to-residual alignment framework for
few-shot generalist anomaly detection.  Unlike existing CLIP-based methods that perform multimodal alignment in the raw feature space, we shift this alignment entirely into a unified residual domain, where both visual and text modalities are represented as residual vectors relative to their normal references. Built on this insight, our framework
extends the conventional dual-branch paradigm with a novel residual branch
that performs symmetric residual-to-residual alignment, and
constrains all learnable optimization within the residual domain to focus on
relative anomaly deviations rather than class-specific absolute features.
Experiments on industrial anomaly detection benchmarks demonstrate that
Res\(^2\)CLIP achieves strong performance on unseen
categories under both training-free and fine-tuning settings, with
particularly strong cross-domain generalization.



{\small
\bibliographystyle{abbrvnat}
\bibliography{ref}
}


\appendix

\section{Theoretical Derivations}
\label{app:derivation}

\paragraph{Notation.}
Throughout this appendix we adopt the following conventions consistent with
the main paper. The hat symbol \(\hat{\,\cdot\,}\) denotes \(\ell_2\)
normalization, so for any vector \(\mathbf{v}\) we write
\(\hat{\mathbf{v}} = \mathbf{v}/\|\mathbf{v}\|_2\). The bracket
\(\langle\,\cdot\,,\,\cdot\,\rangle\) denotes cosine similarity, defined as
\(\langle \mathbf{a}, \mathbf{b} \rangle = \hat{\mathbf{a}} \cdot \hat{\mathbf{b}}\);
when both arguments are unit-norm, this reduces to the standard inner
product. The symbol \(\mathbf{a} \cdot \mathbf{b}\) denotes the unnormalized
inner product along the feature channel dimension. Visual features
\(f^q\), \(f_{\text{ref}}\) and text anchors \(F_t^n\), \(F_t^a\) are CLIP
outputs that are not unit-norm in general; their \(\ell_2\)-normalized
counterparts are \(\hat{f}^q\), \(\hat{f}_{\text{ref}}\), \(\hat{F}_t^n\),
\(\hat{F}_t^a\). The text and visual residuals are constructed from
normalized vectors:
\begin{equation}
\mathbf{R}_t = \hat{F}_t^a - \hat{F}_t^n,
\qquad
\mathbf{R}_v = \hat{f}^q - \hat{f}_{\text{ref}}.
\end{equation}

\subsection{Equivalence of Text Branch Score to Residual Alignment}
\label{app:text_residual}

We show that the softmax-based text anomaly score \(\mathbf{S}_{\text{text}}\)
in Eq.~(2) of the main paper is equivalent to
\(\sigma\!\bigl(\|\mathbf{R}_t\|_2\,\langle f^q, \hat{\mathbf{R}}_t \rangle\bigr)\)
in \cref{eq:text_score2residual}, where \(\hat{\mathbf{R}}_t = \mathbf{R}_t / \|\mathbf{R}_t\|_2\)
is the unit-direction of the text residual.

\paragraph{Step 1: Softmax to sigmoid form.}
Starting from ~\cref{eq:text_score}:
\begin{equation}
\mathbf{S}_{\text{text}}
= \frac{\exp(\langle f^q, F_t^a \rangle)}
       {\exp(\langle f^q, F_t^n \rangle) + \exp(\langle f^q, F_t^a \rangle)}.
\end{equation}
Dividing both numerator and denominator by \(\exp(\langle f^q, F_t^a \rangle)\):
\begin{equation}
\mathbf{S}_{\text{text}}
= \frac{1}{1 + \exp\!\bigl(\langle f^q, F_t^n \rangle - \langle f^q, F_t^a \rangle\bigr)}
= \sigma\!\bigl(\langle f^q, F_t^a \rangle - \langle f^q, F_t^n \rangle\bigr),
\end{equation}
where \(\sigma(x) = 1/(1+e^{-x})\).

\paragraph{Step 2: Express the difference of cosine similarities.}
Expanding the cosine similarities by definition:
\begin{equation}
\langle f^q, F_t^a \rangle - \langle f^q, F_t^n \rangle
= \hat{f}^q \cdot \hat{F}_t^a - \hat{f}^q \cdot \hat{F}_t^n
= \hat{f}^q \cdot \bigl(\hat{F}_t^a - \hat{F}_t^n\bigr)
= \hat{f}^q \cdot \mathbf{R}_t.
\end{equation}

\paragraph{Step 3: Decompose the residual into magnitude and direction.}
Writing \(\mathbf{R}_t = \|\mathbf{R}_t\|_2\,\hat{\mathbf{R}}_t\):
\begin{equation}
\hat{f}^q \cdot \mathbf{R}_t
= \|\mathbf{R}_t\|_2\,\bigl(\hat{f}^q \cdot \hat{\mathbf{R}}_t\bigr)
= \|\mathbf{R}_t\|_2\,\langle f^q, \hat{\mathbf{R}}_t \rangle,
\end{equation}
where the last equality follows because \(\hat{\mathbf{R}}_t\) is unit-norm.
Combining with Step 1:
\begin{equation}
\mathbf{S}_{\text{text}}
= \sigma\!\bigl(\|\mathbf{R}_t\|_2\,\langle f^q, \hat{\mathbf{R}}_t \rangle\bigr).
\end{equation}

\paragraph{Remark.}
The factor \(\|\mathbf{R}_t\|_2\) is a fixed positive scalar independent of
the query \(f^q\). It acts as a temperature scaling on the sigmoid input but
does not affect the relative ranking of anomaly scores across patches. The
decision signal is therefore determined by
\(\langle f^q, \hat{\mathbf{R}}_t \rangle\), the cosine similarity between
the visual feature and the unit-direction of the text residual.

\subsection{Equivalence of Visual Branch Score to Residual Magnitude}
\label{app:visual_residual}

We show that \(\mathbf{S}_{\text{vis}} = 1 - \langle f^q, f_{\text{ref}} \rangle\)
in ~\cref{eq:vis_score} of the main paper is equivalent to
\(\tfrac{1}{2}\|\mathbf{R}_v\|_2^2\) in Eq.~(5).

By definition of cosine similarity,
\(\langle f^q, f_{\text{ref}} \rangle = \hat{f}^q \cdot \hat{f}_{\text{ref}}\).
Recall that \(\mathbf{R}_v = \hat{f}^q - \hat{f}_{\text{ref}}\). Expanding
the squared norm:
\begin{equation}
\begin{aligned}
\|\mathbf{R}_v\|_2^2
&= \|\hat{f}^q - \hat{f}_{\text{ref}}\|_2^2 \\
&= \|\hat{f}^q\|_2^2 - 2\,\hat{f}^q \cdot \hat{f}_{\text{ref}} + \|\hat{f}_{\text{ref}}\|_2^2 \\
&= 1 - 2\,\langle f^q, f_{\text{ref}} \rangle + 1 \\
&= 2\bigl(1 - \langle f^q, f_{\text{ref}} \rangle\bigr) = 2\,\mathbf{S}_{\text{vis}}.
\end{aligned}
\end{equation}
where we used \(\|\hat{f}^q\|_2 = \|\hat{f}_{\text{ref}}\|_2 = 1\). Therefore
\(\mathbf{S}_{\text{vis}} = \tfrac{1}{2}\|\mathbf{R}_v\|_2^2\).

\paragraph{Remark.}
This equivalence shows that the visual branch exploits only the magnitude
\(\|\mathbf{R}_v\|_2\) and discards the directional information of the
visual residual. The vector \(\mathbf{R}_v\) lives in a \(D\)-dimensional
space, and its direction encodes which semantic dimensions deviate from the
normal reference. Collapsing this structure into a scalar magnitude is a
notable information loss, which motivates the explicit use of the full
residual vector in our residual branch.

\subsection{Derivation of Symmetric Residual Alignment}
\label{app:symmetric_alignment}

We derive the equivalence in ~\cref{eq:proj} of the main paper, showing that
calibrating the text-branch alignment with a normal reference is equivalent
to the projection \(\mathbf{R}_v \cdot \hat{\mathbf{R}}_t\).

\paragraph{Step 1: Normal baseline calibration.}
The text-branch alignment between a feature and the text residual is
\(\langle f, \mathbf{R}_t \rangle\). Even when the query is normal, this
alignment is non-zero due to region-type baselines in \(\hat{f}^q\). A
natural calibration subtracts the normal reference's alignment from the
query's:
\begin{equation}
\langle f^q, \mathbf{R}_t \rangle - \langle f_{\text{ref}}, \mathbf{R}_t \rangle.
\end{equation}

\paragraph{Step 2: Express in dot-product form.}
By definition of cosine similarity in our notation
\(\langle \mathbf{a}, \mathbf{b} \rangle = \hat{\mathbf{a}} \cdot \hat{\mathbf{b}}\):
\begin{equation}
\langle f^q, \mathbf{R}_t \rangle - \langle f_{\text{ref}}, \mathbf{R}_t \rangle
= \hat{f}^q \cdot \hat{\mathbf{R}}_t - \hat{f}_{\text{ref}} \cdot \hat{\mathbf{R}}_t.
\end{equation}

\paragraph{Step 3: Apply linearity.}
By linearity of the inner product:
\begin{equation}
\hat{f}^q \cdot \hat{\mathbf{R}}_t - \hat{f}_{\text{ref}} \cdot \hat{\mathbf{R}}_t
= \bigl(\hat{f}^q - \hat{f}_{\text{ref}}\bigr) \cdot \hat{\mathbf{R}}_t
= \mathbf{R}_v \cdot \hat{\mathbf{R}}_t,
\end{equation}
where we used the definition \(\mathbf{R}_v = \hat{f}^q - \hat{f}_{\text{ref}}\).
Combining with Step 2:
\begin{equation}
\langle f^q, \mathbf{R}_t \rangle - \langle f_{\text{ref}}, \mathbf{R}_t \rangle
= \mathbf{R}_v \cdot \hat{\mathbf{R}}_t.
\end{equation}
The calibrated text-branch alignment therefore reduces exactly to \(\mathbf{R}_v \cdot \hat{\mathbf{R}}_t\), the projection of the visual residual onto the unit-direction of the text residual. 

\paragraph{Remark.}
Calibrating the absolute-space text similarity with a normal reference
reduces algebraically to a residual-to-residual projection. This identity
simultaneously addresses two limitations of the absolute-space alignment.
First, the region-type baseline in \(\hat{f}^q\) is removed because the
calibration replaces \(\hat{f}^q\) with the differential \(\mathbf{R}_v\).
Second, the directional information of \(\mathbf{R}_v\) is preserved through
the projection \(\mathbf{R}_v \cdot \hat{\mathbf{R}}_t\), in contrast to
the squared-magnitude score \(\|\mathbf{R}_v\|_2^2\) used by the visual
branch, which discards direction entirely.

\subsection{Why Projection Instead of Cosine Similarity?}
\label{app:projection_vs_cosine}

In the residual branch, we use the projection
\(\mathbf{R}_v \cdot \hat{\mathbf{R}}_t\) rather than cosine similarity
\(\langle \mathbf{R}_v, \hat{\mathbf{R}}_t \rangle\). We discuss the
numerical reason here.

\paragraph{Issue with cosine similarity.}
By definition, the cosine similarity between \(\mathbf{R}_v\) and
\(\hat{\mathbf{R}}_t\) normalizes by the magnitude of \(\mathbf{R}_v\)
(\(\hat{\mathbf{R}}_t\) is already unit-norm):
\begin{equation}
\langle \mathbf{R}_v, \hat{\mathbf{R}}_t \rangle
= \frac{\mathbf{R}_v \cdot \hat{\mathbf{R}}_t}{\|\mathbf{R}_v\|_2}.
\end{equation}
In normal regions, \(\|\mathbf{R}_v\|_2\) is close to zero because the query feature closely matches the retrieved reference. The subtraction has already cancelled out the shared semantic content, so the remaining residual carries little directional information and consists mostly of retrieval noise. Computing cosine similarity on such a near-zero vector forces a meaningless direction onto unit length, turning pure noise into a spurious anomaly score and producing false positives in normal regions.

\paragraph{Property of projection.}
The projection preserves magnitude and avoids small-denominator
amplification:
\begin{equation}
\mathbf{R}_v \cdot \hat{\mathbf{R}}_t \to 0
\quad \text{as} \quad \|\mathbf{R}_v\|_2 \to 0.
\end{equation}
Near-zero residual magnitude naturally yields a near-zero anomaly score.
For anomalous regions where \(\|\mathbf{R}_v\|_2\) is large, the projection
reflects both how far the feature deviates and how well that deviation
aligns with the text-defined anomaly direction.

\paragraph{Remark.}
The magnitude of the visual residual itself carries an anomaly signal,
which cosine similarity would discard by normalization. Projection
preserves this signal and is therefore more appropriate for the residual
branch.

\section{Text Prompt Design}
\label{app:prompt}

This section provides full details of the text prompts used to construct the
normal and anomalous text anchor features \(F_t^n\) and \(F_t^a\).

\subsection{Prompt Construction}

Each prompt is constructed by combining three components:
\begin{itemize}\setlength{\itemsep}{0pt}
    \item A state word \(\texttt{[state]}\) describing the normal or anomalous
    condition, such as \texttt{flawless} or \texttt{damaged}.
    \item A generic noun \(\texttt{[noun]}\) referring to the specific category or a generic object term.
    \item A template \(\mathcal{T}(\cdot)\) that wraps the state-noun phrase into
    a natural sentence.
\end{itemize}
A complete prompt is formed as
\(\mathcal{T}(\texttt{[state]}\ \texttt{[noun]})\). For example,
\(\mathcal{T}(\cdot) = \texttt{a photo of a } \cdot\) with
\(\texttt{[state]} = \texttt{flawless}\) and \(\texttt{[noun]} = \texttt{object}\)
yields the prompt \texttt{a photo of a flawless object}.

\subsection{State Words}

We use small fixed sets of state words for the normal and anomalous conditions:
\begin{itemize}\setlength{\itemsep}{0pt}
    \item Normal: \(\{\text{empty},\, \texttt{perfect},\, \texttt{flawless},\,
    \texttt{clean},\, \texttt{undamaged}\}\). The empty state produces a noun-only
    prompt that captures the neutral semantic baseline.
    \item Anomalous: \(\{\texttt{damaged}\}\).
\end{itemize}

\subsection{Generic Noun}

Following prior work~\cite{zhou2023anomalyclip}, we use a single generic noun
\(\texttt{[noun]} = \texttt{object}\) for all classes.

\subsection{Templates}

We use the following \(15\) templates, following common practice in CLIP-based
zero-shot classification:
\begin{itemize}\setlength{\itemsep}{0pt}
    \item \texttt{a photo of a \{\}.}
    \item \texttt{a photo of the \{\}.}
    \item \texttt{a cropped photo of a \{\}.}
    \item \texttt{a cropped photo of the \{\}.}
    \item \texttt{a close-up photo of a \{\}.}
    \item \texttt{a close-up photo of the \{\}.}
    \item \texttt{the \{\} in the image.}
    \item \texttt{the \{\} in the scene.}
    \item \texttt{an image of a \{\}.}
    \item \texttt{an image of the \{\}.}
    \item \texttt{a rendering of a \{\}.}
    \item \texttt{a photo of one \{\}.}
    \item \texttt{showing the \{\}.}
    \item \texttt{showing a \{\}.}
    \item \texttt{itap of a \{\}.}
\end{itemize}

\subsection{Anchor Computation}

Let \(\mathcal{T}_{\text{tmpl}}\) denote the set of templates listed above, and
let \(\mathcal{W}^n\) and \(\mathcal{W}^a\) denote the sets of normal and
anomalous state words, respectively. The full prompt sets are:
\begin{equation}
\mathcal{P}_t^n = \bigl\{ \mathcal{T}(\texttt{[state]}\ \texttt{[noun]})
    \,:\, \texttt{[state]} \in \mathcal{W}^n,\ \mathcal{T} \in \mathcal{T}_{\text{tmpl}} \bigr\},
\end{equation}
\begin{equation}
\mathcal{P}_t^a = \bigl\{ \mathcal{T}(\texttt{[state]}\ \texttt{[noun]})
    \,:\, \texttt{[state]} \in \mathcal{W}^a,\ \mathcal{T} \in \mathcal{T}_{\text{tmpl}} \bigr\},
\end{equation}
where \(\texttt{[noun]} = \texttt{object}\). Each prompt is encoded by the CLIP
text encoder \(\mathcal{E}_t\), and all encodings within a set are averaged to
yield the final anchor features:
\begin{equation}
F_t^n = \frac{1}{|\mathcal{P}_t^n|} \sum_{T \in \mathcal{P}_t^n} \mathcal{E}_t(T),
\quad
F_t^a = \frac{1}{|\mathcal{P}_t^a|} \sum_{T \in \mathcal{P}_t^a} \mathcal{E}_t(T).
\end{equation}

\paragraph{Remark.}
The same prompt sets and templates are used across all datasets. No per-dataset
prompt tuning is performed.

\section{Sparse Adaptive Matching}
\label{app:sparsemax}

To aggregate the memory bank \(\mathcal{B}_l\) into a query-aligned reference
feature, we adopt the sparse retrieval mechanism introduced in
~\cite{qu2025dictas}, which applies the Sparsemax
operator~\cite{martins2016softmax} to the penalized similarity matrix
\(\mathcal{S}_l \in \mathbb{R}^{N \times (K \cdot N)}\) defined in the visual
branch.

\paragraph{Sparsemax operator.}
For an input vector \(z \in \mathbb{R}^M\), Sparsemax projects \(z\) onto the
probability simplex \(\Delta^{M-1}\) under the squared-error objective:
\begin{equation}
\operatorname{Sparsemax}(z) = \arg\min_{p \in \Delta^{M-1}} \|p - z\|_2^2.
\end{equation}
The solution admits a closed form. After shifting \(z\) by its maximum entry
for numerical stability, the procedure proceeds as follows:
\begin{enumerate}\setlength{\itemsep}{0pt}
    \item Sort the entries of \(z\) in descending order:
    \(z_{(1)} \geq z_{(2)} \geq \cdots \geq z_{(M)}\).
    \item Identify the support size
    \(k(z) = \max\!\bigl\{m \in [M] \,:\, 1 + m\,z_{(m)} > \sum_{j=1}^{m} z_{(j)} \bigr\}\).
    \item Compute the threshold
    \(\tau(z) = \bigl(\sum_{j=1}^{k(z)} z_{(j)} - 1\bigr) / k(z)\).
    \item Output the sparse weights
    \(\operatorname{Sparsemax}(z)_i = \max\!\bigl(0,\, z_i - \tau(z)\bigr)\).
\end{enumerate}
Entries with \(z_i \leq \tau(z)\) are mapped to exactly zero, while the
remaining entries form a probability distribution that sums to 1. 

\paragraph{Application to reference retrieval.}
Applying the operator above row-wise to \(\mathcal{S}_l\) yields a sparse
weight matrix \(W_l \in \mathbb{R}^{N \times (K \cdot N)}\), where each query
row contains a small number of non-zero weights summing to 1. The spatially
aligned reference feature is obtained by weighted aggregation:
\begin{equation}
F_{\text{ref},l} = W_l \, \mathcal{B}_l \in \mathbb{R}^{N \times D}.
\end{equation}

\paragraph{Remark.}
Unlike Softmax-based aggregation, which assigns non-zero weight to every
candidate in \(\mathcal{B}_l\), Sparsemax automatically selects a
data-dependent number of relevant candidates and discards the rest, removing
the need to tune a fixed top-\(k\) hyperparameter. This property is the main
practical reason we adopt this operator: the support size adapts to the
similarity distribution of each query patch, and no per-dataset selection of
\(k\) is required.

\section{Details of Loss Functions}
\label{app:loss}

This section provides the full forms of the loss components summarized in
\cref{sec:optimization} of the main paper.

\subsection{Text Branch Loss}
\label{app:loss_text}

The global and local case-specific penalty losses for the text branch are:
\begin{equation}
\mathcal{L}_{\text{text}}^{\text{global}}
= (1 - \mathbf{y}^{gt}) \max\!\bigl(0,\, \langle f_v^q, \widetilde{\mathbf{R}}_t \rangle\bigr)
+ \mathbf{y}^{gt} \max\!\bigl(0,\, 1 - \langle f_v^q, \widetilde{\mathbf{R}}_t \rangle\bigr),
\end{equation}
\begin{equation}
\mathcal{L}_{\text{text}}^{\text{local}}
= \sum_{l=1}^{L} \!\left(
    \frac{1}{N_0} \!\sum_{\mathbf{M}^{gt}=0}\!
      \max\!\bigl(0,\, \langle F_{v,l}^q, \widetilde{\mathbf{R}}_t \rangle\bigr)
  + \frac{1}{N_1} \!\sum_{\mathbf{M}^{gt}=1}\!
      \max\!\bigl(0,\, 1 - \langle F_{v,l}^q, \widetilde{\mathbf{R}}_t \rangle\bigr)
  \right),
\end{equation}
where \(N_0\) and \(N_1\) are the numbers of normal and anomalous pixels.

\paragraph{Remark.}

The penalty for normal samples is one-sided: once \(\langle f, \widetilde{\mathbf{R}}_t \rangle\) falls below \(0\), the term contributes no gradient. This conservative behavior helps preserve CLIP's pre-trained feature geometry, since normal features are not pushed toward extreme values such as \(-1\) and are allowed to retain their natural diversity within the half-space orthogonal or opposite to the anomaly direction. For anomalous samples, the cosine similarity is upper-bounded by \(1\), so the loss term remains active throughout training and continuously drives \(\langle f, \widetilde{\mathbf{R}}_t \rangle\) toward perfect alignment. The cosine-distance regularizer \(1 - \langle \widetilde{\mathbf{R}}_t,, \mathbf{R}_t \rangle\) further constrains \(\widetilde{\mathbf{R}}_t\) to remain close to the direction of the original residual, preventing the adapter from drifting away from CLIP's semantic manifold.

\subsection{Residual Branch Loss}
\label{app:loss_res}

\paragraph{Hinge-style base functions.}
The base hinge functions used in the alternating residual branch objective are:
\begin{equation}
\ell_{\text{res}}^{\text{global}}(P_{\text{cls}})
= (1 - \mathbf{y}^{gt}) \max\!\bigl(0,\, P_{\text{cls}}\bigr)
+ \mathbf{y}^{gt} \max\!\bigl(0,\, \|\mathbf{R}_{v,\text{cls}}\|_2 - P_{\text{cls}}\bigr),
\end{equation}
\begin{equation}
\ell_{\text{res}}^{\text{local}}(P_l)
= \frac{1}{N_0} \!\sum_{\mathbf{M}^{gt}=0}\!
    \max\!\bigl(0,\, P_l\bigr)
+ \frac{1}{N_1} \!\sum_{\mathbf{M}^{gt}=1}\!
    \max\!\bigl(0,\, \|\mathbf{R}_{v,l}\|_2 - P_l\bigr),
\end{equation}
where \(P_{\text{cls}}\) and \(P_l\) are placeholder projection terms that are
substituted with the appropriate stop-gradient expressions depending on which
adapter is being updated.

\paragraph{Remark on the upper target \(\|\mathbf{R}_v\|_2\).}
The projection of an adapted visual residual
\(\widetilde{\mathbf{R}}_v^{\text{res}}\) onto the normalized text residual
direction \(\hat{\widetilde{\mathbf{R}}}_t^{\text{res}}\) admits the standard
geometric decomposition
\begin{equation}
\widetilde{\mathbf{R}}_v^{\text{res}} \cdot \hat{\widetilde{\mathbf{R}}}_t^{\text{res}}
= \|\widetilde{\mathbf{R}}_v^{\text{res}}\|_2 \cdot \cos\theta,
\end{equation}
where \(\theta\) is the angle between the two residuals. The maximum is
attained at \(\cos\theta = 1\), giving an upper bound of
\(\|\widetilde{\mathbf{R}}_v^{\text{res}}\|_2 \cdot 1
= \|\widetilde{\mathbf{R}}_v^{\text{res}}\|_2\). To avoid letting the adapter
trivially inflate this bound by enlarging
\(\|\widetilde{\mathbf{R}}_v^{\text{res}}\|_2\), we anchor the target to the
raw physical residual magnitude \(\|\mathbf{R}_v\|_2\) instead. The hinge term
\(\max(0, \|\mathbf{R}_v\|_2 - P)\) therefore encourages anomalous projections
to reach the geometric maximum achievable when the adapted visual residual
keeps its original magnitude and aligns perfectly with the text residual
direction. Normal projections are simultaneously pushed below \(0\),
establishing a two-sided margin with the safe interval between the two
thresholds left unpenalized.

\paragraph{Regularization terms.}
The regularization terms for the two modalities are:
\begin{equation}
\mathcal{L}_{\text{reg}}^{t}
= 1 - \langle \widetilde{\mathbf{R}}_t^{\text{res}},\, \mathbf{R}_t \rangle,
\end{equation}
\begin{equation}
\mathcal{L}_{\text{reg}}^{v}
= \bigl|\|\widetilde{\mathbf{R}}_{v,\text{cls}}^{\text{res}}\|_2
        - \|\mathbf{R}_{v,\text{cls}}\|_2\bigr|
+ \sum_{l=1}^{L} \frac{1}{N}
    \bigl|\|\widetilde{\mathbf{R}}_{v,l}^{\text{res}}\|_2
          - \|\mathbf{R}_{v,l}\|_2\bigr|.
\end{equation}

\paragraph{Remark on regularizing the visual magnitude.}
The two regularization terms target different properties of the adapted residuals because the two modalities play different roles in the projection score. The text residual encodes a semantic anomaly direction, so its adapted version is constrained to remain aligned with the original residual via the cosine-distance term \(\mathcal{L}_{\text{reg}}^t\). The visual residual requires a different treatment. Since the projection score factorizes as \(\|\widetilde{\mathbf{R}}_v^{\text{res}}\|_2 \cdot \cos\theta\), the alignment objective can in principle be satisfied through both angular and magnitude adjustments. However, the angular component is naturally bounded by \(\cos\theta \leq 1\), whereas the magnitude has no intrinsic upper limit, making it the easier optimization direction. Without regularization, the adapter would tend to favor magnitude inflation over angular adjustment, leaving the directional alignment under-trained. The \(\ell_1\) magnitude penalty \(\mathcal{L}_{\text{reg}}^{v}\) balances the two by anchoring \(\|\widetilde{\mathbf{R}}_v^{\text{res}}\|_2\) close to the raw magnitude \(\|\mathbf{R}_v\|_2\), so that both components contribute meaningfully to the alignment.

\section{Adapter Architecture}
\label{app:adapter}

This section provides the architectural details of the adapters used in the
fine-tuning setting. In the training-free setting, all adapters degenerate to
the identity mapping. We use \(D = 768\) (CLIP ViT-L/14) and a bottleneck
dimension of \(D/4 = 192\) throughout. All adapters are initialized so that
the output equals the input at the start of training, by setting the final
up-projection to zero.

\subsection{Adapters for Global Features}

The text residual adapters \(\mathcal{A}_t\) and \(\mathcal{A}_t^{\text{res}}\),
together with the global visual residual adapter
\(\mathcal{A}_{v,\text{cls}}^{\text{res}}\), all operate on \(D\)-dimensional
vectors and share the same architecture. For an input residual
\(\mathbf{R} \in \mathbb{R}^{D}\), each adapter applies:
\begin{equation}
\mathcal{A}(\mathbf{R})
= \mathbf{R}
+ \operatorname{MLP}\!\bigl(\operatorname{LN}(\mathbf{R})\bigr),
\end{equation}
where \(\operatorname{LN}\) denotes LayerNorm. The MLP consists of a linear
layer projecting from \(D\) to \(D/4\), a GELU activation, and a second
linear layer projecting from \(D/4\) back to \(D\). The output linear layer
is zero-initialized, so the adapter starts as an identity mapping.

\subsection{Adapters for Local Visual Features}

The local visual residual adapters \(\{\mathcal{A}_{v,l}\}_{l=1}^L\) and
\(\{\mathcal{A}_{v,l}^{\text{res}}\}_{l=1}^L\) operate on patch-level
residual maps of shape \(N \times D\) and are implemented as multi-scale
convolutional adapters. For an input residual map
\(\mathbf{R}_{v,l} \in \mathbb{R}^{N \times D}\), the adapter proceeds in
four stages:

\paragraph{Stage 1: Pre-normalization and channel reduction.}
The input is normalized with LayerNorm and linearly projected from \(D\) to
\(D/4\) channels, followed by a ReLU activation. The resulting sequence is
reshaped into a 2D spatial map of shape \((D/4) \times \sqrt{N} \times
\sqrt{N}\) to enable convolutional operations.

\paragraph{Stage 2: Multi-scale spatial convolution.}
Two depthwise convolutions are applied in parallel to the 2D feature map:
\begin{itemize}\setlength{\itemsep}{0pt}
    \item A \(3{\times}3\) depthwise convolution with dilation 1, capturing
    fine-grained local context.
    \item A \(3{\times}3\) depthwise convolution with dilation 2 (effective
    receptive field \(5{\times}5\)), capturing coarser spatial structure.
\end{itemize}
The pre-convolution feature map and the outputs of both branches are
concatenated along the channel dimension, yielding a map of \(3 \times D/4\)
channels, which is then flattened back to a sequence of shape
\(N \times 3D/4\).

\paragraph{Stage 3: Internal normalization and channel fusion.}
A second LayerNorm is applied to balance the magnitudes of the three
concatenated branches. A linear layer then fuses the \(3D/4\) channels back
to \(D/4\), followed by a ReLU activation.


\paragraph{Stage 4: Up-projection and residual connection.}
A zero-initialized linear layer projects the output of Stage 3 from \(D/4\)
back to \(D\), and the result is added to the original input via a residual
connection:
\begin{equation}
\mathcal{A}_{v,l}(\mathbf{R}_{v,l})
= \mathbf{R}_{v,l} + \operatorname{Linear}_{\text{up}}(\mathbf{Z}_l),
\end{equation}
where \(\mathbf{Z}_l \in \mathbb{R}^{N \times D/4}\) denotes the output of
Stage 3. A separate adapter is trained for each layer
\(l \in \{1, \ldots, L\}\) to address layer-specific noise patterns
independently.

\paragraph{Remark.}
The two adapter families are designed for different signal types. Global
residuals are single vectors with no spatial structure, so a simple MLP
bottleneck suffices. Local visual residuals carry per-patch retrieval noise
that is spatially correlated, so a multi-scale depthwise convolution is used
to capture and suppress this noise.
Parameters are not shared across layers \(l\) or across the visual and
residual branches, because the visual branch enforces a magnitude constraint
on \(\widetilde{\mathbf{R}}_{v,l}\) while the residual branch optimizes its
projection onto the text residual direction. Sharing across these objectives
produces conflicting gradients.

\section{Details of Experimental Setup}
\label{app:exp}

\subsection{Details of Datasets}

~\cref{tab:datasets} summarizes the statistics of all datasets used in
our evaluation. For each dataset, we report the number of object categories,
the number of normal and anomalous test images.

  \begin{table}[h]
\centering
\caption{Statistics of evaluation datasets. \#Normal and \#Anomalous refer
to the number of test images in each split.}
\label{tab:datasets}
\resizebox{0.85\linewidth}{!}{
\begin{tabular}{ l c c c c}
\toprule
 Dataset & \#Categories & \#Normal & \#Anomalous & Aux. Training \\
\midrule

 MVTecAD~\cite{bergmann2019mvtec}  & 15 & 467 & 1258 & VisA \\
 VisA~\cite{zou2022spot}           & 12 & 962 & 1200 & MVTecAD \\
 BTAD~\cite{mishra2021vt}          &  3 & 451 & 290 & MVTecAD \\
  MPDD~\cite{jezek2021deep}          &  6 & 176 & 282 & MVTecAD \\
  DTD-Synthetic~\cite{aota2023zero}        &  12 & 357 & 947 & MVTecAD \\
\bottomrule
\end{tabular}
}
\end{table}

\subsection{More Implementation Details}
\label{app:implementation}
 
\paragraph{Feature extraction.}
We use CLIP ViT-L/14@336px pretrained by OpenAI as the backbone, with V-V
attention applied to the last 20 transformer layers of the visual encoder.
All input images are resized to \(518 \times 518\). The text branch uses
patch features from layer 24 only, in both the training-free and fine-tuning
settings. The visual and residual branches extract features from layers
\(\{6, 12, 18, 24\}\) (\(L = 4\)) in the fine-tuning setting. In the
training-free setting, the residual branch also uses only layer 24, as
deeper features yield stronger cross-domain semantic alignment without
adaptation.
 
\paragraph{Training configuration.}
Adapters are trained with the Adam optimizer using \(\beta_1 = 0.5\),
\(\beta_2 = 0.999\). The local visual adapters use a learning rate of
\(5 \times 10^{-4}\) and the text and global visual adapters and text adapters use
\(1 \times 10^{-4}\). A cosine annealing schedule with one epoch of warmup
is applied. 
We train for 10 epochs with a batch size of 16, applying gradient clipping with a maximum norm of 1.0 to all adapter groups. All experiments are conducted on a single NVIDIA RTX 3090 (24 GB).
The visual and text residual adapters in the residual branch are updated in alternating
epochs. In even-numbered epochs, visual adapters are updated while
the text-side adapter is frozen via \(\operatorname{sg}(\cdot)\); in
odd-numbered epochs, the roles are reversed. The adapters of text and visual branch are updated continuously in every epoch as their losses do not
involve crossmodal coupling.
During training, for each query image we randomly sample one normal image from the same category as the visual prompt, i.e., all training is conducted under a 1-shot setting. The obtained model generalizes directly to larger-shot evaluation without any modification.
 
\paragraph{Inference configuration.}
At inference, the few-shot visual prompt set \(\mathcal{P}_v\) is fixed for
all test images of the same class. 
Anomaly maps are smoothed with a Gaussian kernel (\(\sigma = 4\), kernel
size \(5 \times 5\)). 
The three branches are fused
via weighted summation. The text branch is assigned a lower weight at the
pixel level, as its coarse semantic maps tend to introduce false positives
in fine-grained localization. The residual branch weight differs between
settings to account for the increase in score magnitude after adapter
optimization. 
Concretely, the pixel-level weights are
\(\lambda_{\text{text}}{:}\lambda_{\text{vis}}{:}\lambda_{\text{res}} =
0.1{:}1{:}1\) in the training-free setting and
\(0.1{:}1{:}0.1\) in the fine-tuning setting; the image-level weights are
\(1{:}1{:}1\) and \(1{:}1{:}0.1\) respectively.
The same weights are used
uniformly across all datasets and shots without any per-dataset tuning.
Since I-AUC, P-AUC, P-AP, and PRO are rank-invariant under positive scaling,
the weights need not sum to any specific value.

\paragraph{Evaluation protocol.}
All few-shot experiments are repeated with \(k \in \{1, 2, 4\}\) shots. For
each \(k\), the same set of normal reference images is used across all
compared methods to ensure a fair comparison. Each metric is averaged over three different runs. 
To comprehensively evaluate the model across different aspects of anomaly perception, we report diverse metrics. Specifically, image-level metrics (I-AUC, I-AP, I-F1) assess the global classification performance, where I-AP and I-F1 are strictly included to provide robust evaluation under image-level class imbalance. Pixel-level metrics (P-AUC, P-AP, P-F1) measure the precision of fine-grained spatial localization. Given the extreme numerical imbalance between anomalous and normal pixels, P-AP and P-F1 serve as highly stringent indicators that heavily penalize spatial mismatch and background noise. Furthermore, the Per-Region Overlap (PRO) evaluates localization at the connected-component level, ensuring the assessment is not disproportionately dominated by large anomalous regions and accurately reflects sensitivity to subtle structural defects.

\section{Additional Experimental Results}
\label{app:full_results}

\subsection{Detailed Quantitative Metrics} \label{app:full}
\label{app:sub_detailed_metrics}
\cref{tab:mvtec_full_appendix}, ~\cref{tab:visa_full_appendix}, ~\cref{tab:btad_full_appendix}, ~\cref{tab:mpdd_full_appendix}, and ~\cref{tab:dtd_full_appendix}  provides the complete metric set on datasets MVTecAD~\cite{bergmann2019mvtec}, VisA~\cite{zou2022spot}, BTAD~\cite{mishra2021vt}, MPDD~\cite{jezek2021deep}, and DTD-Synthetic~\cite{aota2023zero} including I-AP,
I-F1, P-F1 for all datasets and all shot counts.
 
\begin{table*}
  \caption{Full quantitative results on MVTecAD~\cite{bergmann2019mvtec} across 1, 2, and 4 shots. All metrics are reported as mean $\pm$ standard deviation over three random runs. The best results are highlighted in \textbf{bold} and the second-best results are \underline{underlined}.}
  \label{tab:mvtec_full_appendix}
  \centering
  \resizebox{\linewidth}{!}{
  \begin{tabular}{c|l|ccccccc}
    \toprule
    Shots & Method & I-AUC & I-AP & I-F1 & P-AUC & P-AP & P-F1 & PRO \\
    \midrule
    \multirow{7}{*}{1} 
    & WinCLIP+~\cite{jeong2023winclip} & 93.1$_{\pm 0.4}$ & 96.6$_{\pm 0.1}$ & 93.7$_{\pm 0.2}$ & 93.6$_{\pm 0.1}$ & 38.5$_{\pm 0.4}$ & 42.7$_{\pm 0.4}$ & 84.1$_{\pm 0.3}$ \\
    & APRIL-GAN~\cite{chen2023zero} & 92.2$_{\pm 0.1}$ & 96.0$_{\pm 0.1}$ & 92.6$_{\pm 0.2}$ & 95.3$_{\pm 0.1}$ & 52.5$_{\pm 0.4}$ & 55.0$_{\pm 0.3}$ & 90.8$_{\pm 0.2}$ \\
    & AnomalyCLIP+~\cite{zhou2023anomalyclip} & 93.4$_{\pm 0.4}$ & 97.1$_{\pm 0.3}$ & 93.7$_{\pm 0.2}$ & 94.8$_{\pm 0.2}$ & 46.2$_{\pm 0.5}$ & 49.1$_{\pm 0.2}$ & 89.5$_{\pm 0.3}$ \\
    & ReMP-AD~\cite{ma2025remp} & 95.2$_{\pm 0.4}$ & 97.5$_{\pm 0.4}$ & 94.4$_{\pm 0.2}$ & 95.5$_{\pm 0.1}$ & \underline{57.3}$_{\pm 0.9}$ & \underline{57.3}$_{\pm 0.8}$ & \underline{91.6}$_{\pm 0.4}$ \\
    & AdaptCLIP~\cite{gao2026adaptclip} & 95.0$_{\pm 0.6}$ & 97.6$_{\pm 0.4}$ & 95.2$_{\pm 0.3}$ & 94.2$_{\pm 0.1}$ & 53.9$_{\pm 1.3}$ & 53.8$_{\pm 1.0}$ & 89.7$_{\pm 0.3}$ \\
    \cmidrule{2-9}
    & Res\(^2\)CLIP\(^*\) (Ours) & \textbf{96.1}$_{\pm 0.3}$ & \textbf{98.4}$_{\pm 0.1}$ & \textbf{95.6}$_{\pm 0.1}$ & \underline{96.2}$_{\pm 0.2}$ & 50.8$_{\pm 0.8}$ & 54.6$_{\pm 0.6}$ & 91.1$_{\pm 0.2}$ \\
    &Res\(^2\)CLIP\(^\dagger\) (Ours) & \underline{96.0}$_{\pm 0.3}$ & \underline{98.3}$_{\pm 0.1}$ & \underline{95.6}$_{\pm 0.2}$ & \textbf{96.6}$_{\pm 0.2}$ & \textbf{57.9}$_{\pm 0.5}$ & \textbf{57.8}$_{\pm 0.4}$ & \textbf{91.7}$_{\pm 0.3}$ \\
    \midrule
    \multirow{7}{*}{2}
    & WinCLIP+~\cite{jeong2023winclip} & 94.7$_{\pm 0.2}$ & 97.3$_{\pm 0.0}$ & 94.6$_{\pm 0.2}$ & 93.9$_{\pm 0.1}$ & 40.0$_{\pm 0.7}$ & 43.9$_{\pm 0.4}$ & 84.6$_{\pm 0.3}$ \\
    & APRIL-GAN~\cite{chen2023zero} & 92.4$_{\pm 0.3}$ & 96.1$_{\pm 0.1}$ & 92.5$_{\pm 0.2}$ & 95.6$_{\pm 0.1}$ & 53.5$_{\pm 0.2}$ & 55.8$_{\pm 0.2}$ & 91.3$_{\pm 0.1}$ \\
    & AnomalyCLIP+~\cite{zhou2023anomalyclip} & 93.7$_{\pm 0.2}$ & 97.2$_{\pm 0.2}$ & 93.6$_{\pm 0.1}$ & 95.0$_{\pm 0.1}$ & 47.7$_{\pm 0.4}$ & 50.1$_{\pm 0.5}$ & 90.2$_{\pm 0.4}$ \\
    & ReMP-AD~\cite{ma2025remp} & 95.2$_{\pm 0.2}$ & 97.4$_{\pm 0.1}$ & 94.4$_{\pm 0.2}$ & 95.5$_{\pm 0.1}$ & \underline{57.2}$_{\pm 0.1}$ & \underline{57.5}$_{\pm 0.2}$ & 91.9$_{\pm 0.2}$ \\
    & AdaptCLIP~\cite{gao2026adaptclip} & 95.7$_{\pm 0.1}$ & 97.8$_{\pm 0.1}$ & 95.4$_{\pm 0.1}$ & 94.5$_{\pm 0.1}$ & 55.5$_{\pm 0.2}$ & 55.3$_{\pm 0.1}$ & 90.3$_{\pm 0.2}$ \\
    \cmidrule{2-9}
    & Res\(^2\)CLIP\(^*\) (Ours) & \underline{96.3}$_{\pm 0.1}$ & \underline{98.3}$_{\pm 0.1}$ & \underline{95.9}$_{\pm 0.2}$ & \underline{96.7}$_{\pm 0.0}$ & 53.3$_{\pm 0.7}$ & 56.6$_{\pm 0.7}$ & \underline{92.0}$_{\pm 0.1}$ \\
    & Res\(^2\)CLIP\(^\dagger\) (Ours) & \textbf{96.9}$_{\pm 0.4}$ & \textbf{98.5}$_{\pm 0.2}$ & \textbf{96.1}$_{\pm 0.6}$ & \textbf{97.1}$_{\pm 0.1}$ & \textbf{60.6}$_{\pm 0.7}$ & \textbf{60.3}$_{\pm 0.9}$ & \textbf{92.7}$_{\pm 0.1}$ \\
    \midrule
    \multirow{7}{*}{4}
    & WinCLIP+~\cite{jeong2023winclip} & 95.1$_{\pm 0.2}$ & 97.5$_{\pm 0.1}$ & 95.1$_{\pm 0.2}$ & 94.3$_{\pm 0.1}$ & 42.0$_{\pm 0.2}$ & 45.7$_{\pm 0.2}$ & 85.5$_{\pm 0.5}$ \\
    & APRIL-GAN~\cite{chen2023zero} & 92.7$_{\pm 0.1}$ & 96.1$_{\pm 0.1}$ & 92.8$_{\pm 0.1}$ & 96.0$_{\pm 0.1}$ & 54.8$_{\pm 0.1}$ & 57.1$_{\pm 0.2}$ & 91.9$_{\pm 0.1}$ \\
    & AnomalyCLIP+~\cite{zhou2023anomalyclip} & 93.8$_{\pm 0.2}$ & 97.3$_{\pm 0.2}$ & 93.8$_{\pm 0.1}$ & 95.2$_{\pm 0.1}$ & 48.6$_{\pm 0.2}$ & 50.9$_{\pm 0.3}$ & 90.7$_{\pm 0.1}$ \\
    & ReMP-AD~\cite{ma2025remp} & 95.8$_{\pm 0.2}$ & 97.7$_{\pm 0.2}$ & 94.7$_{\pm 0.1}$ & 95.7$_{\pm 0.0}$ & \underline{58.0}$_{\pm 0.2}$ & 58.3$_{\pm 0.1}$ & 92.3$_{\pm 0.1}$ \\
    & AdaptCLIP~\cite{gao2026adaptclip} & 96.8$_{\pm 0.2}$ & 98.5$_{\pm 0.0}$ & 96.2$_{\pm 0.2}$ & 94.8$_{\pm 0.1}$ & 57.8$_{\pm 0.1}$ & 57.5$_{\pm 0.1}$ & 91.1$_{\pm 0.0}$ \\
    \cmidrule{2-9}
    & Res\(^2\)CLIP\(^*\) (Ours) & \underline{97.5}$_{\pm 0.2}$ & \underline{98.9}$_{\pm 0.1}$ & \underline{96.8}$_{\pm 0.3}$ & \underline{97.1}$_{\pm 0.0}$ & 55.3$_{\pm 0.3}$ & \underline{58.5}$_{\pm 0.4}$ & \underline{92.8}$_{\pm 0.1}$ \\
    & Res\(^2\)CLIP\(^\dagger\) (Ours) & \textbf{97.9}$_{\pm 0.2}$ & \textbf{99.0}$_{\pm 0.1}$ & \textbf{96.8}$_{\pm 0.1}$ & \textbf{97.4}$_{\pm 0.1}$ & \textbf{62.5}$_{\pm 0.4}$ & \textbf{62.2}$_{\pm 0.4}$ & \textbf{93.5}$_{\pm 0.1}$ \\
    \bottomrule
  \end{tabular}
  }
\end{table*}

\begin{table*}
  \caption{Full quantitative results on VisA~\cite{zou2022spot} across 1, 2, and 4 shots. All metrics are reported as mean $\pm$ standard deviation over three random runs. The best results are highlighted in \textbf{bold} and the second-best results are \underline{underlined}.}
  \label{tab:visa_full_appendix}
  \centering
  \resizebox{\linewidth}{!}{
  \begin{tabular}{c|l|ccccccc}
    \toprule
    Shots & Method & I-AUC & I-AP & I-F1 & P-AUC & P-AP & P-F1 & PRO \\
    \midrule
    \multirow{7}{*}{1} 
    & WinCLIP+~\cite{jeong2023winclip} & 82.9$_{\pm 0.1}$ & 84.5$_{\pm 0.5}$ & 82.1$_{\pm 0.2}$ & 94.8$_{\pm 0.1}$ & 15.5$_{\pm 0.2}$ & 22.6$_{\pm 0.1}$ & 80.2$_{\pm 0.7}$ \\
    & APRIL-GAN~\cite{chen2023zero} & \underline{91.6}$_{\pm 0.4}$ & \underline{93.6}$_{\pm 0.3}$ & \underline{87.1}$_{\pm 0.4}$ & 96.1$_{\pm 0.0}$ & 31.0$_{\pm 0.3}$ & 38.5$_{\pm 0.2}$ & 90.1$_{\pm 0.1}$ \\
    & AnomalyCLIP+~\cite{zhou2023anomalyclip} & 80.0$_{\pm 0.8}$ & 82.2$_{\pm 0.4}$ & 81.0$_{\pm 0.2}$ & 96.5$_{\pm 0.1}$ & 28.2$_{\pm 0.3}$ & 37.1$_{\pm 0.5}$ & 89.6$_{\pm 0.2}$ \\
    & ReMP-AD~\cite{ma2025remp} & \textbf{92.1}$_{\pm 0.6}$ & \textbf{94.0}$_{\pm 0.3}$ & \textbf{87.8}$_{\pm 0.5}$ & 96.9$_{\pm 0.1}$ & 35.3$_{\pm 0.4}$ & 42.1$_{\pm 0.5}$ & \underline{91.8}$_{\pm 0.1}$ \\
    & AdaptCLIP~\cite{gao2026adaptclip} & 91.2$_{\pm 0.9}$ & 93.0$_{\pm 0.7}$ & 87.0$_{\pm 1.0}$ & \underline{97.0}$_{\pm 0.1}$ & \textbf{40.0}$_{\pm 0.4}$ & \underline{45.2}$_{\pm 0.4}$ & 91.2$_{\pm 0.1}$ \\
    \cmidrule{2-9}
    & Res\(^2\)CLIP\(^*\) (Ours) & 88.2$_{\pm 0.3}$ & 89.4$_{\pm 0.1}$ & 85.4$_{\pm 0.2}$ & 96.6$_{\pm 0.1}$ & 28.2$_{\pm 0.2}$ & 37.3$_{\pm 0.1}$ & 89.6$_{\pm 0.2}$ \\
    & Res\(^2\)CLIP\(^\dagger\) (Ours) & 89.5$_{\pm 0.5}$ & 90.7$_{\pm 0.1}$ & 86.1$_{\pm 0.3}$ & \textbf{97.6}$_{\pm 0.0}$ & \underline{39.2}$_{\pm 0.3}$ & \textbf{45.3}$_{\pm 0.2}$ & \textbf{92.0}$_{\pm 0.1}$ \\
    \midrule
    \multirow{7}{*}{2}
    & WinCLIP+~\cite{jeong2023winclip} & 83.5$_{\pm 0.4}$ & 84.9$_{\pm 0.6}$ & 82.5$_{\pm 0.3}$ & 95.0$_{\pm 0.1}$ & 16.5$_{\pm 0.6}$ & 23.9$_{\pm 0.7}$ & 80.8$_{\pm 0.3}$ \\
    & APRIL-GAN~\cite{chen2023zero} & 92.3$_{\pm 0.2}$ & \underline{94.2}$_{\pm 0.2}$ & 87.7$_{\pm 0.3}$ & 96.1$_{\pm 0.1}$ & 31.6$_{\pm 0.1}$ & 39.1$_{\pm 0.2}$ & 90.1$_{\pm 0.0}$ \\
    & AnomalyCLIP+~\cite{zhou2023anomalyclip} & 82.2$_{\pm 0.7}$ & 84.6$_{\pm 0.2}$ & 81.1$_{\pm 0.3}$ & 96.4$_{\pm 0.1}$ & 28.8$_{\pm 0.3}$ & 37.6$_{\pm 0.2}$ & 89.4$_{\pm 0.3}$ \\
    & ReMP-AD~\cite{ma2025remp} & \underline{92.6}$_{\pm 0.4}$ & \textbf{94.3}$_{\pm 0.5}$ & \textbf{88.7}$_{\pm 0.5}$ & 96.8$_{\pm 0.1}$ & 35.1$_{\pm 0.2}$ & 42.1$_{\pm 0.2}$ & \underline{91.9}$_{\pm 0.1}$ \\
    & AdaptCLIP~\cite{gao2026adaptclip} & \textbf{92.7}$_{\pm 0.0}$ & 94.1$_{\pm 0.2}$ & \underline{88.2}$_{\pm 0.1}$ & \underline{97.2}$_{\pm 0.1}$ & \textbf{41.1}$_{\pm 0.4}$ & \underline{46.4}$_{\pm 0.5}$ & 91.6$_{\pm 0.4}$ \\
    \cmidrule{2-9}
    & Res\(^2\)CLIP\(^*\) (Ours) & 89.3$_{\pm 0.1}$ & 90.6$_{\pm 0.3}$ & 85.8$_{\pm 0.1}$ & 96.9$_{\pm 0.1}$ & 30.3$_{\pm 1.0}$ & 39.3$_{\pm 0.7}$ & 90.3$_{\pm 0.4}$ \\
    & Res\(^2\)CLIP\(^\dagger\) (Ours) & 90.4$_{\pm 0.4}$ & 91.4$_{\pm 0.1}$ & 86.7$_{\pm 0.4}$ & \textbf{97.8}$_{\pm 0.0}$ & \underline{40.6}$_{\pm 0.7}$ & \textbf{46.7}$_{\pm 0.6}$ & \textbf{92.7}$_{\pm 0.5}$ \\
    \midrule
    \multirow{7}{*}{4}
    & WinCLIP+~\cite{jeong2023winclip} & 84.4$_{\pm 1.0}$ & 85.6$_{\pm 1.6}$ & 82.9$_{\pm 0.9}$ & 95.2$_{\pm 0.1}$ & 17.9$_{\pm 1.2}$ & 25.4$_{\pm 1.1}$ & 81.3$_{\pm 0.2}$ \\
    & APRIL-GAN~\cite{chen2023zero} & 92.8$_{\pm 0.2}$ & \underline{94.6}$_{\pm 0.2}$ & 88.3$_{\pm 0.2}$ & 96.2$_{\pm 0.0}$ & 31.8$_{\pm 0.0}$ & 39.5$_{\pm 0.3}$ & 90.1$_{\pm 0.1}$ \\
    & AnomalyCLIP+~\cite{zhou2023anomalyclip} & 81.7$_{\pm 0.7}$ & 84.0$_{\pm 0.7}$ & 81.3$_{\pm 0.6}$ & 96.5$_{\pm 0.1}$ & 29.3$_{\pm 0.1}$ & 38.0$_{\pm 0.1}$ & 89.7$_{\pm 0.2}$ \\
    & ReMP-AD~\cite{ma2025remp} & \textbf{93.3}$_{\pm 0.3}$ & \textbf{94.8}$_{\pm 0.3}$ & \textbf{89.0}$_{\pm 0.2}$ & 96.9$_{\pm 0.1}$ & 34.8$_{\pm 0.2}$ & 42.5$_{\pm 0.2}$ & 91.9$_{\pm 0.1}$ \\
    & AdaptCLIP~\cite{gao2026adaptclip} & \textbf{93.3}$_{\pm 0.3}$ & 94.5$_{\pm 0.3}$ & \underline{88.8}$_{\pm 0.3}$ & \underline{97.3}$_{\pm 0.1}$ & \textbf{42.0}$_{\pm 0.4}$ & \underline{47.2}$_{\pm 0.3}$ & \underline{92.0}$_{\pm 0.2}$ \\
    \cmidrule{2-9}
    & Res\(^2\)CLIP\(^*\) (Ours) & 90.1$_{\pm 0.4}$ & 90.9$_{\pm 0.6}$ & 86.8$_{\pm 0.5}$ & \underline{97.3}$_{\pm 0.0}$ & 31.4$_{\pm 0.3}$ & 40.5$_{\pm 0.1}$ & 90.8$_{\pm 0.1}$ \\
    & Res\(^2\)CLIP\(^\dagger\) (Ours) & 91.1$_{\pm 0.4}$ & 91.6$_{\pm 0.4}$ & 87.5$_{\pm 0.2}$ & \textbf{98.0}$_{\pm 0.0}$ & \underline{41.0}$_{\pm 0.4}$ & \textbf{47.3}$_{\pm 0.4}$ & \textbf{93.3}$_{\pm 0.1}$ \\
    \bottomrule
  \end{tabular}
  }
\end{table*}

 \begin{table*}
  \caption{Full quantitative results on BTAD~\cite{mishra2021vt} across 1, 2, and 4 shots. All metrics are reported as mean $\pm$ standard deviation over three random runs. The best results are highlighted in \textbf{bold} and the second-best results are \underline{underlined}.}
  \label{tab:btad_full_appendix}
  \centering
  \resizebox{\linewidth}{!}{
  \begin{tabular}{c|l|ccccccc}
    \toprule
    Shots & Method & I-AUC & I-AP & I-F1 & P-AUC & P-AP & P-F1 & PRO \\
    \midrule
    \multirow{7}{*}{1} 
    & WinCLIP+~\cite{jeong2023winclip} & 85.0$_{\pm 3.7}$ & 80.6$_{\pm 4.4}$ & 78.2$_{\pm 3.9}$ & 95.6$_{\pm 0.4}$ & 42.0$_{\pm 1.5}$ & 47.4$_{\pm 1.6}$ & 66.7$_{\pm 1.4}$ \\
    & APRIL-GAN~\cite{chen2023zero} & 91.7$_{\pm 1.2}$ & 94.7$_{\pm 1.5}$ & 91.3$_{\pm 0.9}$ & 94.1$_{\pm 0.2}$ & 50.1$_{\pm 0.5}$ & 53.2$_{\pm 0.5}$ & 78.6$_{\pm 1.1}$ \\
    & AnomalyCLIP+~\cite{zhou2023anomalyclip} & 92.7$_{\pm 0.9}$ & 93.8$_{\pm 1.2}$ & 89.6$_{\pm 1.5}$ & 96.6$_{\pm 0.3}$ & 58.4$_{\pm 0.9}$ & 57.6$_{\pm 0.6}$ & 79.4$_{\pm 0.6}$ \\
    & ReMP-AD~\cite{ma2025remp} & 94.9$_{\pm 0.8}$ & 96.1$_{\pm 1.1}$ & \underline{92.1}$_{\pm 1.5}$ & 95.5$_{\pm 0.3}$ & 53.9$_{\pm 1.0}$ & 56.7$_{\pm 0.7}$ & 82.3$_{\pm 1.3}$ \\
    & AdaptCLIP~\cite{gao2026adaptclip} & 93.1$_{\pm 0.6}$ & \underline{96.2}$_{\pm 1.4}$ & 91.7$_{\pm 2.0}$ & 96.7$_{\pm 0.3}$ & 61.4$_{\pm 0.9}$ & 59.3$_{\pm 0.5}$ & 77.7$_{\pm 0.8}$ \\
    \cmidrule{2-9}
    & Res\(^2\)CLIP\(^*\) (Ours) & \underline{95.5}$_{\pm 0.7}$ & 94.8$_{\pm 2.1}$ & 89.7$_{\pm 1.9}$ & \underline{97.6}$_{\pm 0.3}$ & \underline{63.0}$_{\pm 1.4}$ & \underline{61.3}$_{\pm 1.3}$ & \underline{82.8}$_{\pm 0.7}$ \\
    & Res\(^2\)CLIP\(^\dagger\) (Ours) & \textbf{95.9}$_{\pm 0.2}$ & \textbf{96.8}$_{\pm 1.3}$ & \textbf{92.3}$_{\pm 1.6}$ & \textbf{98.1}$_{\pm 0.2}$ & \textbf{64.2}$_{\pm 2.0}$ & \textbf{61.7}$_{\pm 1.4}$ & \textbf{83.0}$_{\pm 0.7}$ \\
    \midrule
    \multirow{7}{*}{2}
    & WinCLIP+~\cite{jeong2023winclip} & 85.7$_{\pm 0.5}$ & 85.3$_{\pm 1.4}$ & 80.4$_{\pm 1.9}$ & 95.8$_{\pm 0.1}$ & 43.3$_{\pm 1.5}$ & 49.1$_{\pm 1.1}$ & 66.7$_{\pm 0.6}$ \\
    & APRIL-GAN~\cite{chen2023zero} & 91.9$_{\pm 0.7}$ & 95.2$_{\pm 0.7}$ & 91.6$_{\pm 0.3}$ & 94.2$_{\pm 0.1}$ & 50.6$_{\pm 0.4}$ & 53.5$_{\pm 0.3}$ & 78.3$_{\pm 0.7}$ \\
    & AnomalyCLIP+~\cite{zhou2023anomalyclip} & 92.5$_{\pm 0.3}$ & 93.4$_{\pm 0.6}$ & 89.0$_{\pm 1.3}$ & 96.6$_{\pm 0.1}$ & 59.1$_{\pm 0.5}$ & 58.1$_{\pm 0.5}$ & 79.0$_{\pm 0.4}$ \\
    & ReMP-AD~\cite{ma2025remp} & 95.2$_{\pm 0.5}$ & 96.8$_{\pm 0.4}$ & 92.3$_{\pm 0.3}$ & 95.4$_{\pm 0.1}$ & 54.4$_{\pm 0.3}$ & 57.2$_{\pm 0.3}$ & 82.3$_{\pm 0.6}$ \\
    & AdaptCLIP~\cite{gao2026adaptclip} & 92.9$_{\pm 0.7}$ & \underline{96.9}$_{\pm 0.7}$ & \underline{93.3}$_{\pm 1.3}$ & 96.7$_{\pm 0.1}$ & 61.7$_{\pm 0.6}$ & 59.3$_{\pm 0.4}$ & 77.9$_{\pm 0.4}$ \\
    \cmidrule{2-9}
    & Res\(^2\)CLIP\(^*\) (Ours) & \underline{95.7}$_{\pm 0.2}$ & 95.8$_{\pm 0.6}$ & 90.0$_{\pm 0.9}$ & \underline{97.7}$_{\pm 0.1}$ & \underline{64.5}$_{\pm 0.3}$ & \underline{62.3}$_{\pm 0.2}$ & \underline{83.1}$_{\pm 0.3}$ \\
    & Res\(^2\)CLIP\(^\dagger\) (Ours) & \textbf{96.2}$_{\pm 0.3}$ & \textbf{97.8}$_{\pm 0.6}$ & \textbf{93.8}$_{\pm 1.0}$ & \textbf{98.2}$_{\pm 0.1}$ & \textbf{65.7}$_{\pm 1.2}$ & \textbf{62.6}$_{\pm 0.8}$ & \textbf{83.5}$_{\pm 0.5}$ \\
    \midrule
    \multirow{7}{*}{4}
    & WinCLIP+~\cite{jeong2023winclip} & 86.4$_{\pm 1.5}$ & 85.4$_{\pm 3.9}$ & 83.6$_{\pm 1.0}$ & 95.9$_{\pm 0.0}$ & 45.0$_{\pm 0.4}$ & 50.5$_{\pm 0.2}$ & 67.0$_{\pm 0.4}$ \\
    & APRIL-GAN~\cite{chen2023zero} & 91.8$_{\pm 0.6}$ & 94.7$_{\pm 0.4}$ & 91.1$_{\pm 0.3}$ & 94.4$_{\pm 0.2}$ & 51.1$_{\pm 0.1}$ & 53.9$_{\pm 0.2}$ & 78.5$_{\pm 0.8}$ \\
    & AnomalyCLIP+~\cite{zhou2023anomalyclip} & 92.4$_{\pm 0.2}$ & 93.4$_{\pm 0.5}$ & 88.8$_{\pm 0.7}$ & 96.7$_{\pm 0.1}$ & 59.3$_{\pm 0.1}$ & 58.2$_{\pm 0.2}$ & 79.1$_{\pm 0.2}$ \\
    & ReMP-AD~\cite{ma2025remp} & 94.4$_{\pm 0.5}$ & 96.4$_{\pm 0.7}$ & 91.9$_{\pm 1.0}$ & 95.5$_{\pm 0.2}$ & 55.6$_{\pm 0.1}$ & 57.8$_{\pm 0.1}$ & 82.0$_{\pm 0.8}$ \\
    & AdaptCLIP~\cite{gao2026adaptclip} & 93.1$_{\pm 0.1}$ & \underline{96.7}$_{\pm 0.3}$ & \underline{92.6}$_{\pm 0.6}$ & 96.8$_{\pm 0.1}$ & 62.8$_{\pm 0.2}$ & 60.0$_{\pm 0.1}$ & 78.3$_{\pm 0.1}$ \\
    \cmidrule{2-9}
    & Res\(^2\)CLIP\(^*\) (Ours) & \underline{96.4}$_{\pm 0.5}$ & 96.4$_{\pm 0.6}$ & 92.1$_{\pm 1.0}$ & \underline{97.8}$_{\pm 0.1}$ & \underline{65.6}$_{\pm 0.3}$ & \underline{63.0}$_{\pm 0.2}$ & \underline{83.2}$_{\pm 0.1}$ \\
    & Res\(^2\)CLIP\(^\dagger\) (Ours) & \textbf{96.8}$_{\pm 0.5}$ & \textbf{98.3}$_{\pm 0.3}$ & \textbf{95.5}$_{\pm 0.4}$ & \textbf{98.2}$_{\pm 0.0}$ & \textbf{67.1}$_{\pm 0.5}$ & \textbf{63.4}$_{\pm 0.4}$ & \textbf{83.5}$_{\pm 0.2}$ \\
    \bottomrule
  \end{tabular}
  }
\end{table*}

\begin{table*}
  \caption{Full quantitative results on MPDD~\cite{jezek2021deep} across 1, 2, and 4 shots. All metrics are reported as mean $\pm$ standard deviation over three random runs. The best results are highlighted in \textbf{bold} and the second-best results are \underline{underlined}.}
  \label{tab:mpdd_full_appendix}
  \centering
  \resizebox{\linewidth}{!}{
  \begin{tabular}{c|l|ccccccc}
    \toprule
    Shots & Method & I-AUC & I-AP & I-F1 & P-AUC & P-AP & P-F1 & PRO \\
    \midrule
    \multirow{7}{*}{1} 
    & WinCLIP+~\cite{jeong2023winclip} & 69.1$_{\pm 0.8}$ & 73.7$_{\pm 1.0}$ & 81.1$_{\pm 0.4}$ & 96.2$_{\pm 0.3}$ & 30.7$_{\pm 0.6}$ & 32.6$_{\pm 0.9}$ & 88.4$_{\pm 1.2}$ \\
    & APRIL-GAN~\cite{chen2023zero} & 82.6$_{\pm 1.5}$ & 84.9$_{\pm 1.5}$ & 84.4$_{\pm 3.2}$ & 96.7$_{\pm 0.0}$ & 35.6$_{\pm 0.5}$ & 40.5$_{\pm 0.7}$ & 91.2$_{\pm 0.1}$ \\
    & AnomalyCLIP+~\cite{zhou2023anomalyclip} & 79.7$_{\pm 2.1}$ & 80.5$_{\pm 2.3}$ & 82.1$_{\pm 0.9}$ & 97.4$_{\pm 0.1}$ & 36.8$_{\pm 0.7}$ & 41.5$_{\pm 0.6}$ & 93.5$_{\pm 0.3}$ \\
    & ReMP-AD~\cite{ma2025remp} & 82.8$_{\pm 1.5}$ & 85.4$_{\pm 1.4}$ & 83.0$_{\pm 1.0}$ & 97.6$_{\pm 0.1}$ & \textbf{38.9}$_{\pm 0.9}$ & \textbf{42.9}$_{\pm 1.2}$ & 93.7$_{\pm 0.2}$ \\
    & AdaptCLIP~\cite{gao2026adaptclip} & \underline{84.1}$_{\pm 0.3}$ & \underline{85.7}$_{\pm 0.6}$ & \underline{85.1}$_{\pm 1.5}$ & \underline{97.7}$_{\pm 0.1}$ & 34.7$_{\pm 2.0}$ & 37.9$_{\pm 2.0}$ & 93.7$_{\pm 0.3}$ \\
    \cmidrule{2-9}
    & Res\(^2\)CLIP\(^*\) (Ours) & \textbf{86.6}$_{\pm 1.0}$ & \textbf{89.4}$_{\pm 1.1}$ & \textbf{86.1}$_{\pm 1.2}$ & 97.6$_{\pm 0.1}$ & 36.9$_{\pm 0.9}$ & 41.6$_{\pm 0.9}$ & \underline{94.0}$_{\pm 0.3}$ \\
    & Res\(^2\)CLIP\(^\dagger\) (Ours) & 83.6$_{\pm 0.4}$ & 84.2$_{\pm 1.0}$ & 83.9$_{\pm 1.2}$ & \textbf{98.0}$_{\pm 0.0}$ & \underline{37.3}$_{\pm 0.4}$ & \underline{41.7}$_{\pm 0.6}$ & \textbf{94.2}$_{\pm 0.3}$ \\
    \midrule
    \multirow{7}{*}{2}
    & WinCLIP+~\cite{jeong2023winclip} & 70.1$_{\pm 0.7}$ & 74.0$_{\pm 0.9}$ & 81.9$_{\pm 0.6}$ & 95.1$_{\pm 1.0}$ & 31.9$_{\pm 0.6}$ & 33.7$_{\pm 0.6}$ & 89.5$_{\pm 0.3}$ \\
    & APRIL-GAN~\cite{chen2023zero} & 84.6$_{\pm 0.9}$ & 89.2$_{\pm 1.1}$ & 85.3$_{\pm 1.2}$ & 96.7$_{\pm 0.1}$ & 36.9$_{\pm 0.3}$ & 41.5$_{\pm 0.7}$ & 91.3$_{\pm 0.2}$ \\
    & AnomalyCLIP+~\cite{zhou2023anomalyclip} & 78.1$_{\pm 0.9}$ & 82.4$_{\pm 6.8}$ & 82.6$_{\pm 0.6}$ & 97.5$_{\pm 0.1}$ & 37.6$_{\pm 0.5}$ & 42.3$_{\pm 0.6}$ & 94.1$_{\pm 0.1}$ \\
    & ReMP-AD~\cite{ma2025remp} & \underline{87.0}$_{\pm 1.2}$ & \underline{89.5}$_{\pm 0.8}$ & 86.9$_{\pm 0.6}$ & 97.7$_{\pm 0.1}$ & \textbf{41.4}$_{\pm 0.7}$ & \textbf{45.3}$_{\pm 1.0}$ & 93.8$_{\pm 0.2}$ \\
    & AdaptCLIP~\cite{gao2026adaptclip} & 85.4$_{\pm 1.1}$ & 87.1$_{\pm 1.6}$ & 86.2$_{\pm 0.3}$ & \underline{97.8}$_{\pm 0.1}$ & 35.7$_{\pm 0.1}$ & 37.8$_{\pm 0.3}$ & 94.3$_{\pm 0.1}$ \\
    \cmidrule{2-9}
    &Res\(^2\)CLIP\(^*\) (Ours) & \textbf{87.6}$_{\pm 0.7}$ & \textbf{89.9}$_{\pm 0.5}$ & \textbf{87.9}$_{\pm 1.1}$ & 97.7$_{\pm 0.2}$ & 38.3$_{\pm 0.2}$ & \underline{42.8}$_{\pm 0.1}$ & \underline{94.5}$_{\pm 0.2}$ \\
    &Res\(^2\)CLIP\(^\dagger\) (Ours) & 85.8$_{\pm 0.2}$ & 87.5$_{\pm 1.1}$ & \underline{87.3}$_{\pm 0.5}$ & \textbf{98.1}$_{\pm 0.1}$ & \underline{39.0}$_{\pm 0.9}$ & 42.7$_{\pm 0.8}$ & \textbf{94.7}$_{\pm 0.2}$ \\
    \midrule
    \multirow{7}{*}{4}
    & WinCLIP+~\cite{jeong2023winclip} & 71.8$_{\pm 1.3}$ & 75.9$_{\pm 0.9}$ & 82.5$_{\pm 0.2}$ & 94.3$_{\pm 0.2}$ & 32.2$_{\pm 0.3}$ & 33.8$_{\pm 0.3}$ & 88.5$_{\pm 0.6}$ \\
    & APRIL-GAN~\cite{chen2023zero} & 85.9$_{\pm 0.5}$ & 90.1$_{\pm 1.1}$ & 86.4$_{\pm 0.2}$ & 96.7$_{\pm 0.0}$ & 37.3$_{\pm 0.3}$ & 41.6$_{\pm 0.4}$ & 91.1$_{\pm 0.2}$ \\
    & AnomalyCLIP+~\cite{zhou2023anomalyclip} & 78.4$_{\pm 0.1}$ & 79.4$_{\pm 0.5}$ & 82.7$_{\pm 0.5}$ & 97.5$_{\pm 0.1}$ & 38.6$_{\pm 0.3}$ & 43.2$_{\pm 0.1}$ & 94.2$_{\pm 0.2}$ \\
    & ReMP-AD~\cite{ma2025remp} & \textbf{89.6}$_{\pm 0.3}$ & \textbf{91.8}$_{\pm 0.3}$ & 87.5$_{\pm 0.3}$ & 97.7$_{\pm 0.0}$ & \textbf{43.1}$_{\pm 0.2}$ & \textbf{47.0}$_{\pm 0.6}$ & 93.6$_{\pm 0.1}$ \\
    & AdaptCLIP~\cite{gao2026adaptclip} & 86.3$_{\pm 0.4}$ & 87.8$_{\pm 0.8}$ & 87.6$_{\pm 0.5}$ & \underline{98.1}$_{\pm 0.1}$ & 39.5$_{\pm 1.1}$ & 41.8$_{\pm 0.8}$ & 94.9$_{\pm 0.1}$ \\
    \cmidrule{2-9}
    &Res\(^2\)CLIP\(^*\) (Ours) & \underline{89.0}$_{\pm 0.6}$ & \underline{91.5}$_{\pm 0.4}$ & \textbf{88.6}$_{\pm 0.9}$ & 97.9$_{\pm 0.1}$ & 40.0$_{\pm 1.0}$ & 44.2$_{\pm 0.9}$ & \underline{95.0}$_{\pm 0.2}$ \\
    & Res\(^2\)CLIP\(^\dagger\) (Ours) & 87.3$_{\pm 0.9}$ & 89.4$_{\pm 1.0}$ & \underline{88.0}$_{\pm 0.5}$ & \textbf{98.3}$_{\pm 0.1}$ & \underline{42.6}$_{\pm 1.3}$ & \underline{46.7}$_{\pm 1.3}$ & \textbf{95.4}$_{\pm 0.2}$ \\
    \bottomrule
  \end{tabular}
  }
\end{table*}

\begin{table*}[t]
  \caption{Full quantitative results on DTD-Synthetic~\cite{aota2023zero} across 1, 2, and 4 shots. All metrics are reported as mean $\pm$ standard deviation over three random runs. The best results are highlighted in \textbf{bold} and the second-best results are \underline{underlined}.}
  \label{tab:dtd_full_appendix}
  \centering
  \resizebox{\linewidth}{!}{
  \begin{tabular}{c|l|ccccccc}
    \toprule
    Shots & Method & I-AUC & I-AP & I-F1 & P-AUC & P-AP & P-F1 & PRO \\
    \midrule
    \multirow{7}{*}{1} 
    & WinCLIP+~\cite{jeong2023winclip} & 98.1$_{\pm 0.2}$ & 99.1$_{\pm 0.1}$ & 96.9$_{\pm 0.1}$ & 96.8$_{\pm 0.1}$ & 48.8$_{\pm 0.4}$ & 52.0$_{\pm 0.4}$ & 90.3$_{\pm 0.1}$ \\
    & APRIL-GAN~\cite{chen2023zero} & \underline{98.4}$_{\pm 0.1}$ & \underline{99.4}$_{\pm 0.0}$ & \underline{97.6}$_{\pm 0.1}$ & 96.5$_{\pm 0.1}$ & 76.5$_{\pm 0.1}$ & 73.9$_{\pm 0.1}$ & 92.0$_{\pm 0.2}$ \\
    & AnomalyCLIP+~\cite{zhou2023anomalyclip} & 95.3$_{\pm 0.6}$ & 98.1$_{\pm 0.3}$ & 94.5$_{\pm 0.4}$ & 96.3$_{\pm 0.2}$ & 70.6$_{\pm 0.6}$ & 68.0$_{\pm 0.4}$ & 91.5$_{\pm 0.3}$ \\
    & ReMP-AD~\cite{ma2025remp} & \textbf{99.3}$_{\pm 0.1}$ & \textbf{99.7}$_{\pm 0.1}$ & \textbf{98.7}$_{\pm 0.2}$ & 96.6$_{\pm 0.2}$ & \textbf{79.2}$_{\pm 0.4}$ & \textbf{76.3}$_{\pm 0.3}$ & 92.5$_{\pm 0.4}$ \\
    & AdaptCLIP~\cite{gao2026adaptclip} & 98.0$_{\pm 0.2}$ & 99.3$_{\pm 0.1}$ & 97.4$_{\pm 0.3}$ & \underline{97.3}$_{\pm 0.2}$ & 76.9$_{\pm 0.3}$ & 72.5$_{\pm 0.2}$ & \underline{93.1}$_{\pm 0.2}$ \\
    \cmidrule{2-9}
    & Res\(^2\)CLIP\(^*\) (Ours) & 95.5$_{\pm 0.3}$ & 98.1$_{\pm 0.3}$ & 94.3$_{\pm 0.4}$ & 95.3$_{\pm 0.3}$ & 69.5$_{\pm 1.0}$ & 68.7$_{\pm 0.6}$ & 90.3$_{\pm 0.3}$ \\
    & Res\(^2\)CLIP\(^\dagger\) (Ours) & 97.2$_{\pm 0.1}$ & 99.0$_{\pm 0.1}$ & 96.1$_{\pm 0.3}$ & \textbf{97.4}$_{\pm 0.2}$ & \underline{77.5}$_{\pm 0.4}$ & \underline{74.1}$_{\pm 0.1}$ & \textbf{94.6}$_{\pm 0.9}$ \\
    \midrule
    \multirow{7}{*}{2}
    & WinCLIP+~\cite{jeong2023winclip} & 98.2$_{\pm 0.1}$ & 99.2$_{\pm 0.1}$ & 97.2$_{\pm 0.2}$ & 96.8$_{\pm 0.1}$ & 49.3$_{\pm 0.3}$ & 52.3$_{\pm 0.3}$ & 90.4$_{\pm 0.3}$ \\
    & APRIL-GAN~\cite{chen2023zero} & \underline{98.4}$_{\pm 0.2}$ & \underline{99.4}$_{\pm 0.1}$ & \underline{97.8}$_{\pm 0.2}$ & 96.7$_{\pm 0.0}$ & 76.9$_{\pm 0.1}$ & 74.2$_{\pm 0.0}$ & 92.2$_{\pm 0.1}$ \\
    & AnomalyCLIP+~\cite{zhou2023anomalyclip} & 95.0$_{\pm 0.6}$ & 98.0$_{\pm 0.2}$ & 94.6$_{\pm 0.1}$ & 96.7$_{\pm 0.0}$ & 71.2$_{\pm 0.2}$ & 68.4$_{\pm 0.1}$ & 91.7$_{\pm 0.1}$ \\
    & ReMP-AD~\cite{ma2025remp} & \textbf{99.4}$_{\pm 0.1}$ & \textbf{99.7}$_{\pm 0.0}$ & \textbf{98.8}$_{\pm 0.1}$ & 97.0$_{\pm 0.0}$ & \textbf{80.7}$_{\pm 0.2}$ & \textbf{77.3}$_{\pm 0.1}$ & 93.1$_{\pm 0.1}$ \\
    & AdaptCLIP~\cite{gao2026adaptclip} & 98.2$_{\pm 0.2}$ & 99.3$_{\pm 0.1}$ & 97.5$_{\pm 0.2}$ & \underline{97.7}$_{\pm 0.2}$ & 78.0$_{\pm 0.3}$ & 73.3$_{\pm 0.1}$ & \underline{93.2}$_{\pm 0.2}$ \\
    \cmidrule{2-9}
    & Res\(^2\)CLIP\(^*\) (Ours) & 95.9$_{\pm 0.3}$ & 98.4$_{\pm 0.2}$ & 94.9$_{\pm 0.2}$ & 95.9$_{\pm 0.0}$ & 71.2$_{\pm 0.2}$ & 69.9$_{\pm 0.1}$ & 90.8$_{\pm 0.4}$ \\
    & Res\(^2\)CLIP\(^\dagger\) (Ours) & 97.4$_{\pm 0.1}$ & 99.0$_{\pm 0.1}$ & 96.5$_{\pm 0.2}$ & \textbf{97.7}$_{\pm 0.1}$ & \underline{78.4}$_{\pm 0.2}$ & \underline{74.8}$_{\pm 0.2}$ & \textbf{94.7}$_{\pm 0.1}$ \\
    \midrule
    \multirow{7}{*}{4}
    & WinCLIP+~\cite{jeong2023winclip} & 98.3$_{\pm 0.1}$ & 99.2$_{\pm 0.1}$ & 97.1$_{\pm 0.1}$ & 97.0$_{\pm 0.1}$ & 50.0$_{\pm 0.2}$ & 52.7$_{\pm 0.2}$ & 90.7$_{\pm 0.1}$ \\
    & APRIL-GAN~\cite{chen2023zero} & \underline{98.6}$_{\pm 0.1}$ & \underline{99.5}$_{\pm 0.1}$ & \underline{97.8}$_{\pm 0.2}$ & 96.8$_{\pm 0.1}$ & 77.3$_{\pm 0.1}$ & 74.5$_{\pm 0.1}$ & 92.5$_{\pm 0.1}$ \\
    & AnomalyCLIP+~\cite{zhou2023anomalyclip} & 95.3$_{\pm 0.2}$ & 98.1$_{\pm 0.1}$ & 94.6$_{\pm 0.2}$ & 96.9$_{\pm 0.1}$ & 71.8$_{\pm 0.1}$ & 68.7$_{\pm 0.1}$ & 92.2$_{\pm 0.1}$ \\
    & ReMP-AD~\cite{ma2025remp} & \textbf{99.4}$_{\pm 0.1}$ & \textbf{99.7}$_{\pm 0.1}$ & \textbf{98.9}$_{\pm 0.1}$ & 97.1$_{\pm 0.1}$ & \textbf{81.2}$_{\pm 0.1}$ & \textbf{77.7}$_{\pm 0.1}$ & 93.5$_{\pm 0.1}$ \\
    & AdaptCLIP~\cite{gao2026adaptclip} & 98.2$_{\pm 0.1}$ & 99.3$_{\pm 0.1}$ & 97.7$_{\pm 0.3}$ & \underline{97.8}$_{\pm 0.1}$ & 78.4$_{\pm 0.2}$ & 73.8$_{\pm 0.2}$ & \underline{93.6}$_{\pm 0.1}$ \\
    \cmidrule{2-9}
    & Res\(^2\)CLIP\(^*\) (Ours) & 96.9$_{\pm 0.1}$ & 98.8$_{\pm 0.1}$ & 96.0$_{\pm 0.3}$ & 96.6$_{\pm 0.1}$ & 73.2$_{\pm 0.1}$ & 71.2$_{\pm 0.1}$ & 92.2$_{\pm 0.2}$ \\
    & Res\(^2\)CLIP\(^\dagger\) (Ours) & 97.7$_{\pm 0.1}$ & 99.1$_{\pm 0.1}$ & 96.8$_{\pm 0.2}$ & \textbf{98.0}$_{\pm 0.1}$ & \underline{79.2}$_{\pm 0.2}$ & \underline{75.1}$_{\pm 0.2}$ & \textbf{95.0}$_{\pm 0.1}$ \\
    \bottomrule
  \end{tabular}
  }
\end{table*}

\subsection{Efficiency and Performance Trade-off}
\label{app:sub_efficiency_tradeoff}

\begin{figure}[h]
    \centering
    \includegraphics[width=0.7\linewidth]{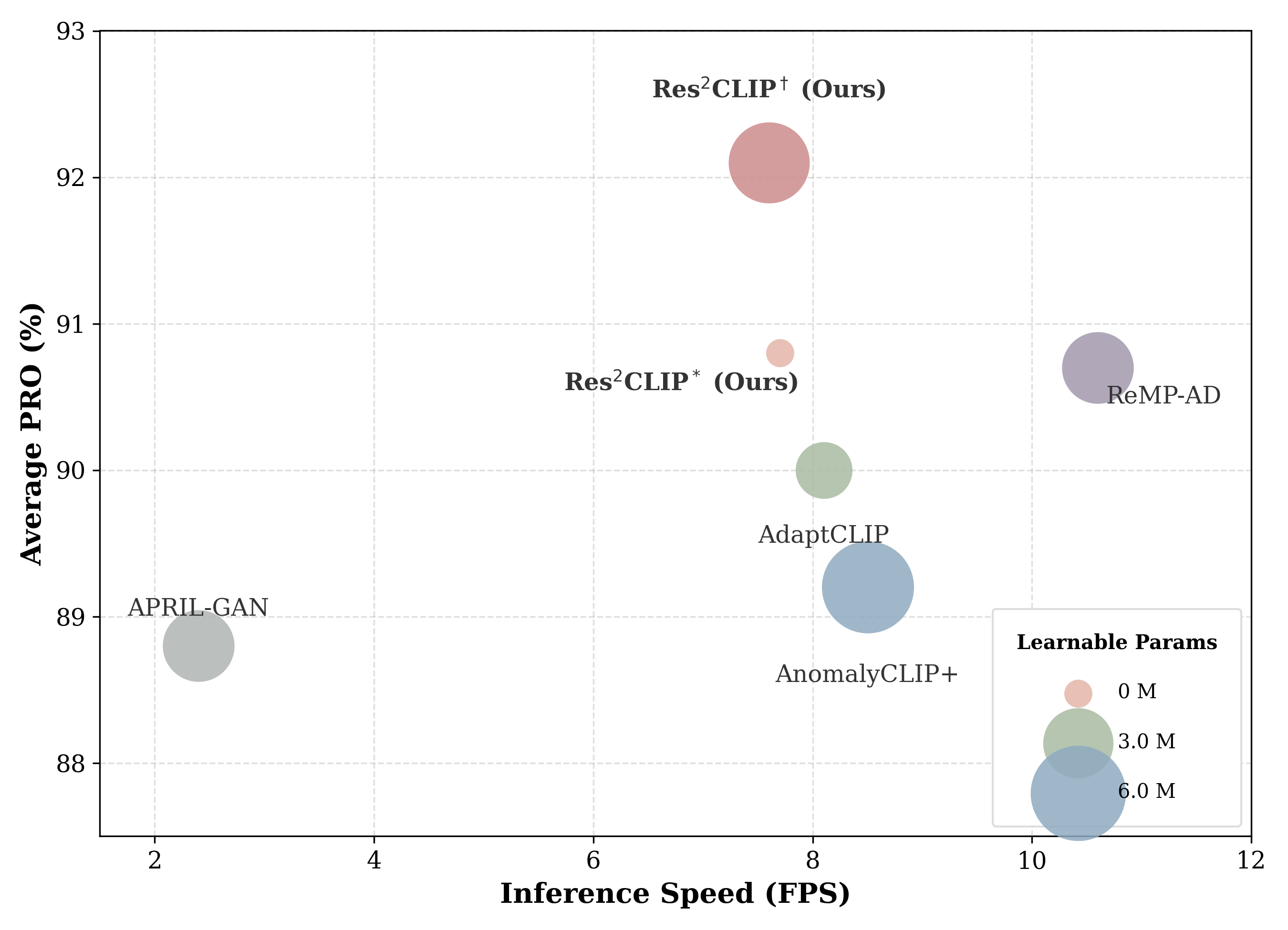}
    \caption{
        Trade-off between inference speed, localization performance, and parameter efficiency. The bubble size denotes the number of extra learnable parameters (e.g., our Res$^2$CLIP$^*$ requires 0M). Our method achieves the optimal balance, delivering state-of-the-art performance with minimal additional parameters and competitive inference speed.
    }
    \label{fig:efficiency_bubble}
\end{figure}

To further demonstrate the practical deployment value of our proposed method, we visualize the trade-off between inference speed, localization performance (PRO), and parameter efficiency under the 4-shot setting. As illustrated in \cref{fig:efficiency_bubble}, the x-axis represents the inference speed (FPS), the y-axis indicates the average pixel-level localization performance (Average PRO), and the bubble size corresponds to the number of extra learnable parameters. 

Notably, our training-free variant Res\(^2\)CLIP\(^*\) achieves highly competitive localization accuracy (90.8\% PRO) with \textbf{zero} additional learnable parameters, demonstrating that the introduction of the residual branch introduces negligible computational overhead. Furthermore, by introducing merely 4.18M parameters, our fine-tuned variant (Res\(^2\)CLIP\(^\dagger\)) achieves state-of-the-art localization precision (92.1\% PRO) with a minimal speed drop (from 7.7 to 7.6 FPS). This comparison demonstrates that our design achieves a highly competitive accuracy‑efficiency trade‑off, balancing strong anomaly detection capability with inference speed suitable for real‑time inspection.

\section{Additional Ablation Studies}
\label{app:ablation}
 
\subsection{Text Anchor Design for Text-Visual Alignment}
 
The text branch computes an anomaly score based on alignment with text
features. We compare three configurations in the training-free setting to
identify which component is effective:
\begin{itemize}
    \item \textit{Normal prompt}: the anomaly score is the \textit{negative}
    cosine similarity between the visual feature and the normal text anchor
    \(F_t^n\), i.e., \(1 - \langle f, \hat{F}_t^n \rangle\). A higher score
    indicates the visual feature is farther from the normal anchor.
    \item \textit{Anomalous prompt}: the anomaly score is the cosine
    similarity between the visual feature and the anomalous text anchor
    \(F_t^a\), i.e., \(\langle f, \hat{F}_t^a \rangle\).
    \item \textit{Residual prompt (ours)}: the anomaly score is the alignment with
    the text residual \(\mathbf{R}_t = \hat{F}_t^a - \hat{F}_t^n\), i.e.,
    \(\langle f, \mathbf{R}_t \rangle\), which is equivalent to using both
    anchors jointly via softmax (as shown in Appendix~\ref{app:text_residual}).
\end{itemize}
 
\begin{table}[h]
\centering
\caption{Ablation on the text anchors in the training-free
setting.}
\label{tab:prompt_ablation}
\resizebox{0.75\linewidth}{!}{
\begin{tabular}{l l ccc}
\toprule
Dataset & Metric & Normal prompt & Anomalous prompt & Residual prompt \\
\midrule
\multirow{3}{*}{MVTecAD}
& I-AUC & 55.0 & 87.3 & 91.5 \\
& P-AP  &  4.7 & 25.3 & 30.9 \\
& PRO   & 27.2 & 70.9 & 83.2 \\
\midrule
\multirow{3}{*}{VisA}
& I-AUC & 54.5 & 76.7 & 82.8 \\
& P-AP  &  0.6 & 11.9 & 16.5 \\
& PRO   & 18.6 & 76.1 & 85.0 \\
\midrule
\multirow{3}{*}{BTAD}
& I-AUC & 56.4 & 77.6 & 91.9 \\
& P-AP  &  7.4 & 27.6 & 43.8 \\
& PRO   & 24.9 & 65.4 & 78.2 \\
\bottomrule
\end{tabular}
}
\end{table}
 
Using only the \textit{Normal prompt} produces near-random performance across all
datasets, because the normal text anchor captures category-level semantics
shared by both normal and anomalous patches rather than an anomaly
direction. Using only the \textit{Anomalous prompt} substantially improves over the
normal-only baseline, confirming that the anomalous anchor carries
discriminative signal. However, the residual formulation \textit{Residual prompt}
consistently outperforms both single-anchor variants, demonstrating that the
\textit{differential direction} between the anomalous and normal anchors but
not either anchor individually, is the effective signal for anomaly
detection. This validates the core motivation of the text residual design.

\subsection{Local Adapter Architecture}
 
\cref{tab:adapter_ablation} compares three architectural choices for the
local visual residual adapters under the fine-tuning setting:
(1)~\textit{Convolution}: our multi-scale depthwise convolutional adapter; (2)~\textit{Linear}: a standard MLP
bottleneck identical to the global adapters; (3)~\textit{Transformer}: a
self-attention block followed by a two-layer MLP.
 
\begin{table}[h]
\centering
\caption{Ablation on the architecture of local adapter in the fine-tuning setting.}
\label{tab:adapter_ablation}
\resizebox{0.65\linewidth}{!}{
\begin{tabular}{l l ccc}
\toprule
Dataset & Metric & Convolution (ours) & Linear & Transformer \\
\midrule
\multirow{3}{*}{MVTecAD}
& I-AUC & \textbf{96.3} & 95.6 & 95.9 \\
& P-AP  & \textbf{57.4} & 56.5 & 55.8 \\
& PRO   & \textbf{91.4} & 91.0 & 90.9 \\
\midrule
\multirow{3}{*}{VisA}
& I-AUC & 89.8 & 90.3 & \textbf{90.6} \\
& P-AP  & \textbf{39.5} & 37.1 & 37.4 \\
& PRO   & \textbf{92.1} & 91.7 & 91.8 \\
\bottomrule
\end{tabular}
}
\end{table}
 
The convolutional adapter achieves the best or competitive performance
across most metrics and datasets, particularly on P-AP. This is consistent
with the design motivation in Appendix~\ref{app:adapter}: local visual
residuals carry spatially correlated retrieval noise that a multi-scale
depthwise convolution is well-suited to suppress, whereas attention-based
and purely linear architectures lack this spatial inductive bias.

\subsection{Selection of Hyperparameter  \(\gamma\)}

\begin{figure}[h]
    \centering
    \includegraphics[width=\linewidth]{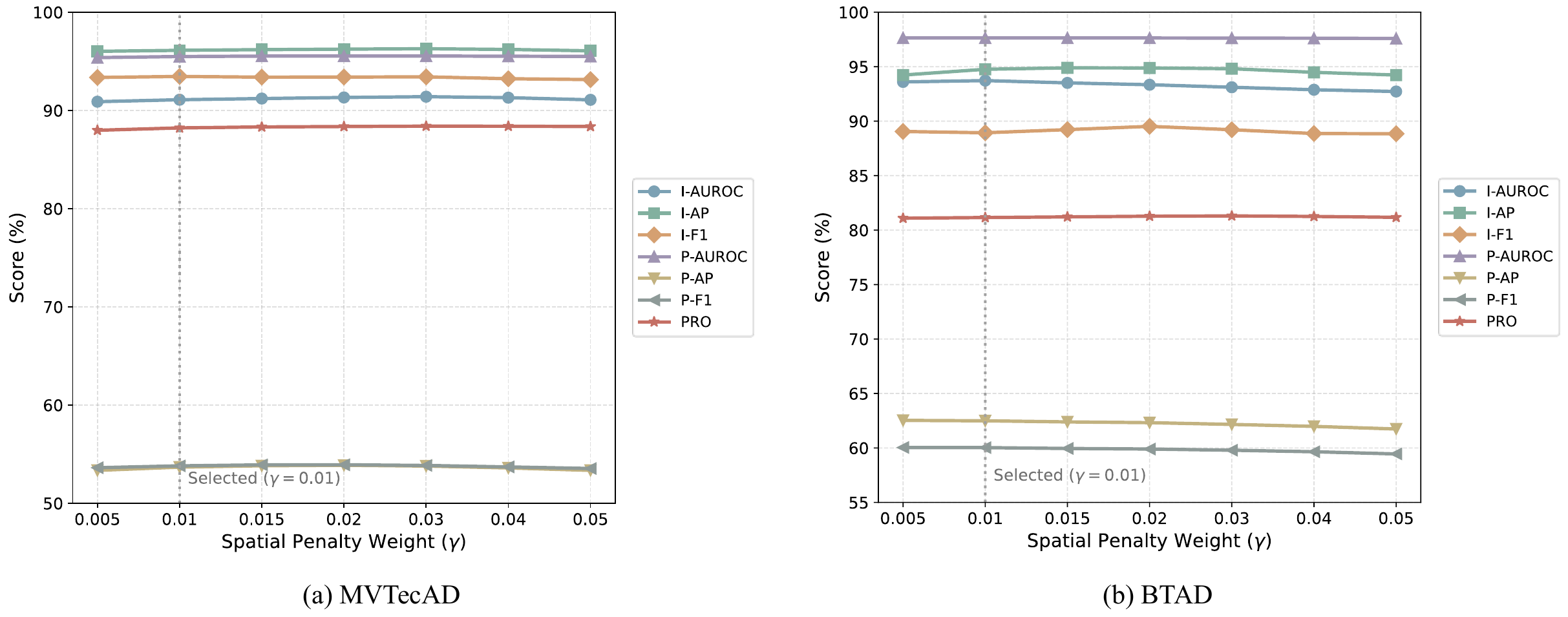} 
    \caption{Ablation on the selection of hyperparameter \(\gamma\) across MVTecAD and BTAD. The performance across all metrics is highly stable, with a consistent optimal sweet spot at \(\gamma = 0.01\). This value is selected as the unified default setting for all experiments.}
    \label{fig:gamma_selection}
\end{figure}

To determine the optimal value for the radial penalty weight \(\gamma\) introduced in ~\cref{sec:visual_branch}, we evaluate the model's performance across varying \(\gamma\) values. ~\cref{fig:gamma_selection} illustrates the performance trends on the MVTecAD and BTAD datasets when \(\gamma\) is varied from \(0.005\) to \(0.05\).

As observed, the performance curves across all metrics remain highly stable on both datasets, demonstrating that \(\gamma\) is not sensitive. Notably, we identify a consistent peak or optimal balance in localization metrics (e.g., P-AP and PRO) around \(\gamma = 0.01\).  Based on these observations, we select \(\gamma = 0.01\) as the unified default setting for all experiments, which avoids the need for any dataset-specific tuning.

\section{Additional Qualitative Visualizations}
\label{app:vis}

\subsection{Residual Feature Distributions.}

To delve deeper into the mechanism of residual representations in multimodal alignment, we conduct comprehensive t-SNE visualizations of the feature space before and after residual extraction.

\paragraph{Decoupling Category-Specific Information.} 

\begin{figure}
    \centering
    \includegraphics[width=0.8\linewidth]{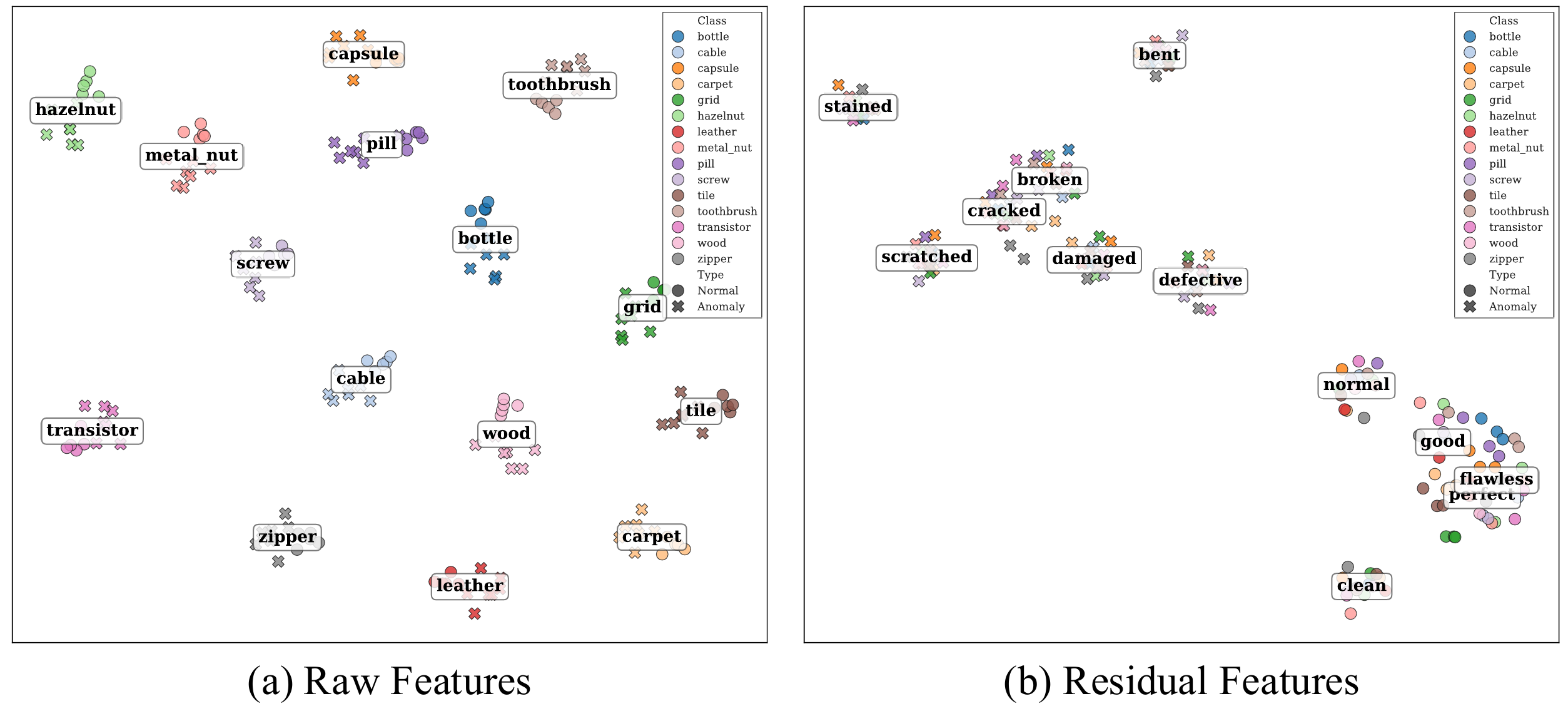}
    \caption{t-SNE visualization of text features on MVTecAD. 
    }
    \label{fig:residual_tsne_text}
\end{figure}

\begin{figure}
    \centering
    \includegraphics[width=0.8\linewidth]{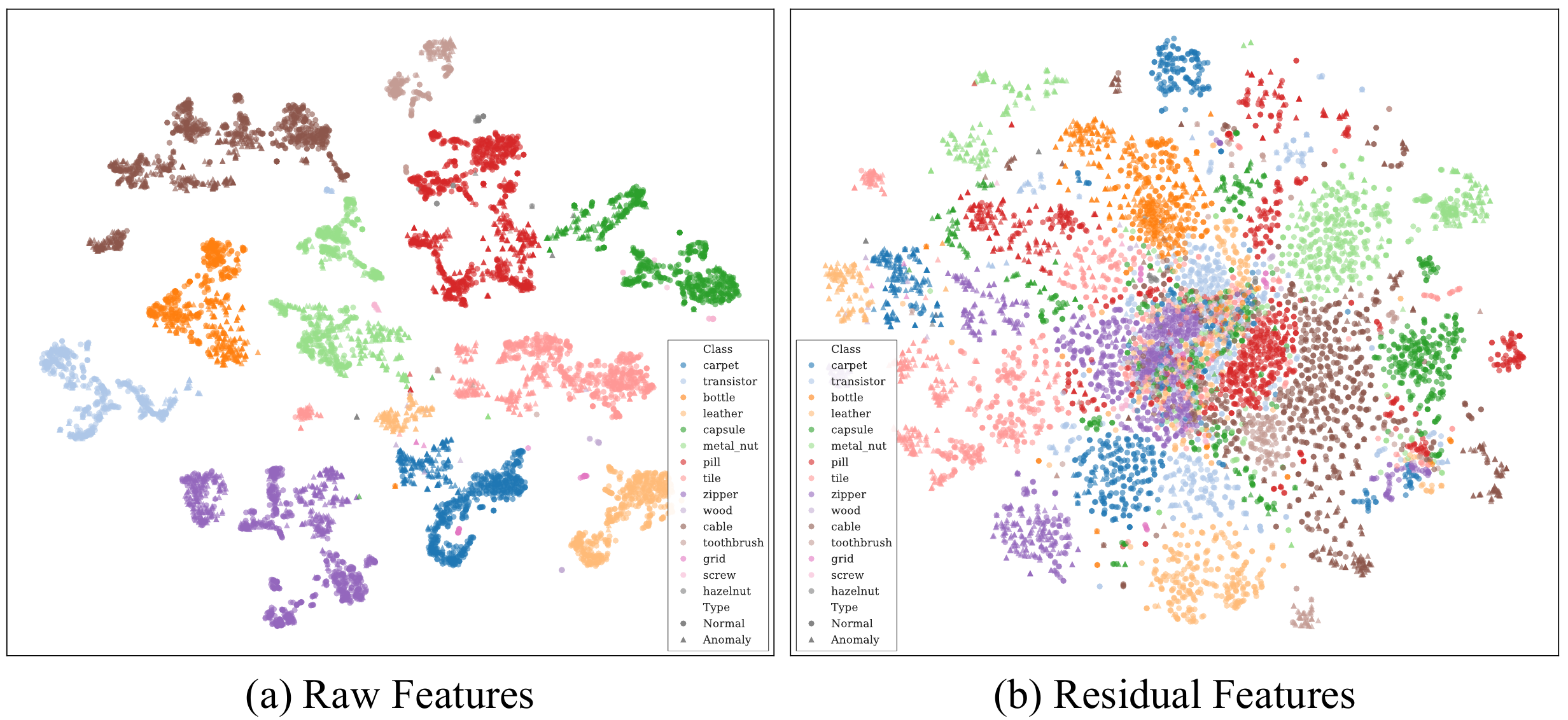}
    \caption{t-SNE visualization of visual features on MVTecAD. 
    }
    \label{fig:residual_tsne_visual}
\end{figure}

To explicitly observe the distribution of different states (both normal and anomalous), the text residual features visualized in this section are specifically computed by subtracting the neutral base feature (i.e., features extracted without any state descriptors) from the state-specific feature. As shown in \cref{fig:residual_tsne_text} and \cref{fig:residual_tsne_visual}, in the original absolute CLIP feature space, the visual and textual features are predominantly governed by object semantics, exhibiting a highly category-dependent clustered distribution (e.g., ``bottle'', ``hazelnut''). Conversely, after shifting to this decoupled residual space, complex category-specific information is significantly suppressed. The feature distribution is no longer clustered by object identity, but is instead strictly dominated by the pure state deviations (e.g., "normal," "damaged"). This intuitively demonstrates that the residual representations effectively mitigate category-specific biases, isolating the core anomaly or normal attributes required for robust detection.

\begin{figure}
    \centering
    \includegraphics[width=0.8\linewidth]{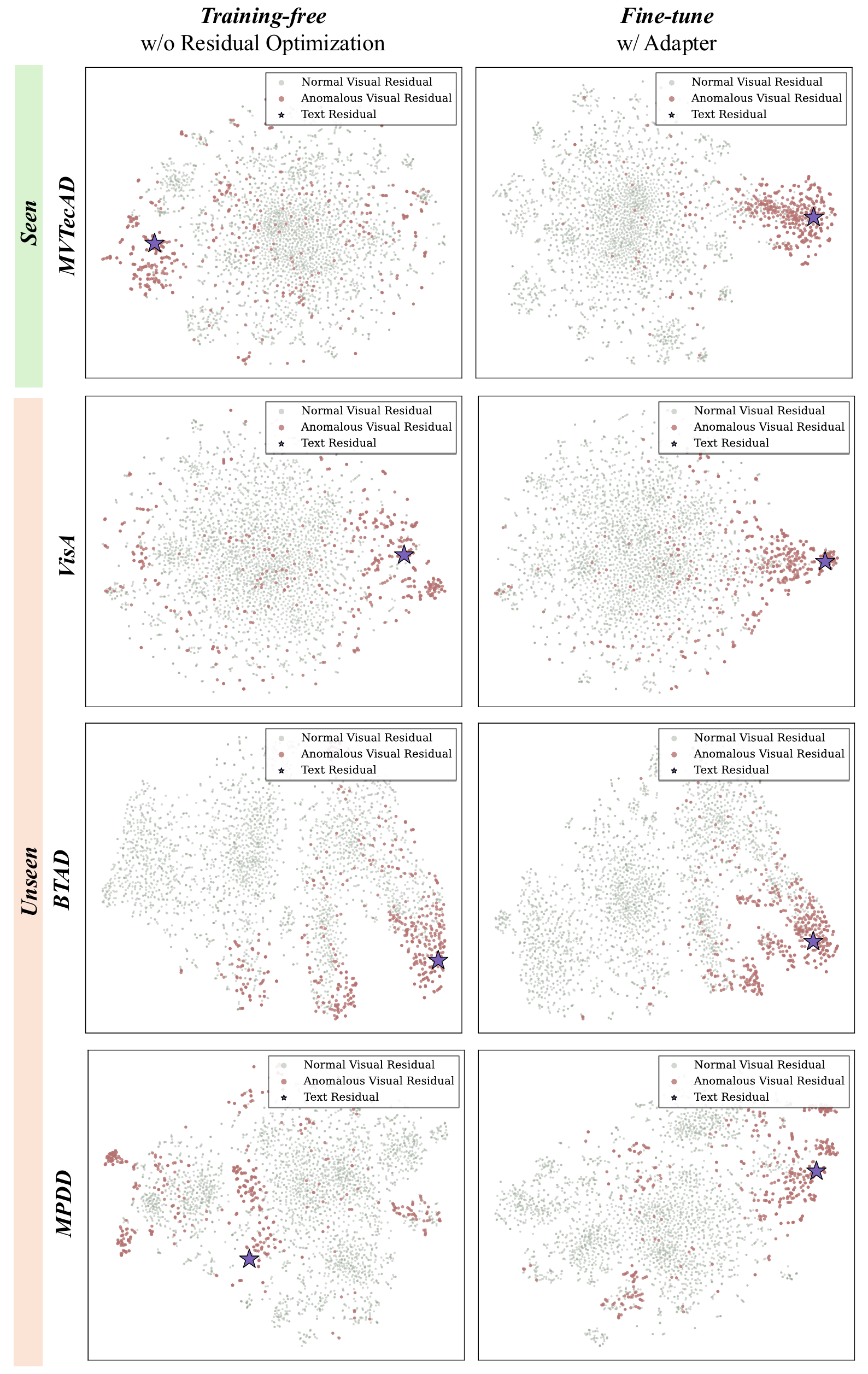}
    \caption{More t-SNE visualizations of residual distributions in the residual branch before and after optimization across multiple datasets. 
    }
    \label{fig:adapter_app}
\end{figure}

\paragraph{Eliminating the Inherent Modality Gap.} 
\cref{fig:adapter_app} further illustrates the relationship between text and visual residuals across multiple datasets. Notably, in the training-free setting (\cref{fig:adapter_app}, left), even without any parameter optimization, the anomalous visual residuals and the text residual anchor already exhibit a clear alignment trend, indicating the absence of a significant modality gap. This empirically validates our mathematical derivations in \cref{sec:residual_branch} and strongly proves that performing text-visual alignment within a unified residual domain naturally bridges the inherent modality gap present in the original CLIP space.

\paragraph{Effectiveness of Residual Optimization.} 
While the quantitative improvements brought by residual optimization are reported in \cref{sec:ablation}, \cref{fig:adapter_app} provides further qualitative evidence. Comparing the left and right panels of \cref{fig:adapter_app}, it is evident that after optimization in the residual domain (w/ Adapter), the anomalous visual residuals are substantially purified and tightly clustered around the text residual anchor. Simultaneously, the margin of separation between the normal visual residuals and the anomalous visual residuals is drastically enlarged. 

 \subsection{Branch-wise Anomaly Maps.}
\begin{figure}
    \centering
    \includegraphics[width=\linewidth]{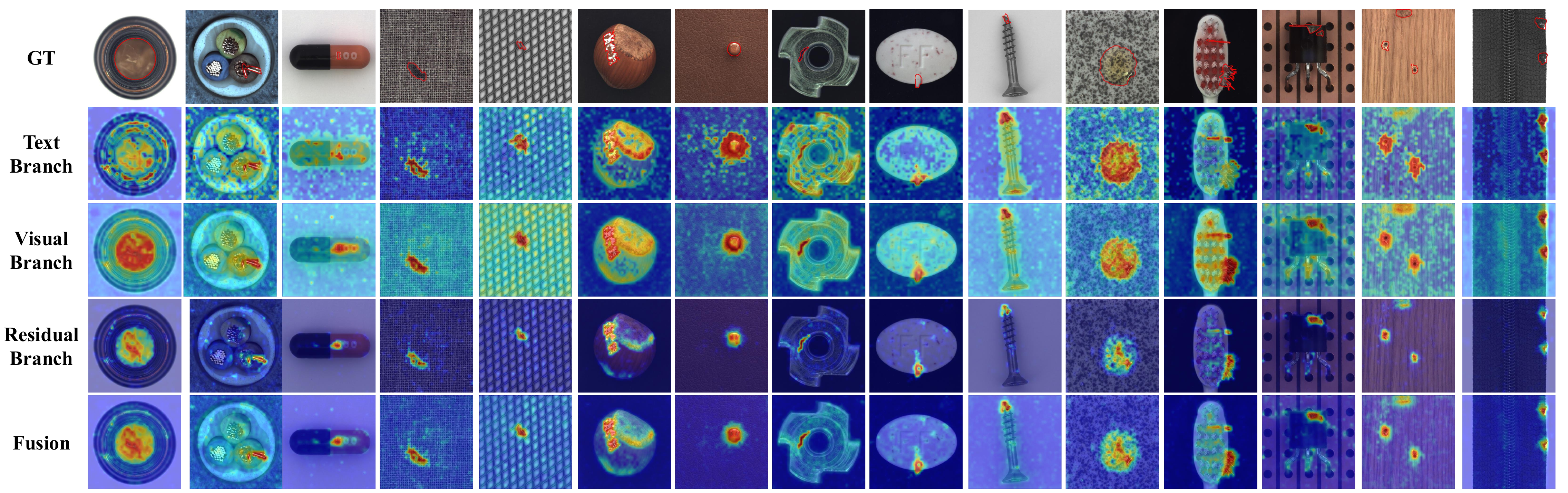}
    \caption{Anomaly map comparison between three branches under the 1-shot setting. 
    }
    \label{fig:branch_vis}
\end{figure}

\cref{fig:branch_vis} compares the anomaly maps from the text branch
(\(\mathbf{M}_{\text{text}}\)), the visual branch
(\(\mathbf{M}_{\text{vis}}\)), the residual branch
(\(\mathbf{M}_{\text{res}}\)), and the fused output \(\mathbf{M}\) on
representative examples. The text branch highlights broad semantic
regions but is coarse; the visual branch captures fine-grained local
deviations but is sensitive to retrieval noise in normal regions; the
residual branch combines the strengths of both by aligning the visual
deviation with the textual anomaly direction. The fused output achieves
tight and complete localization.

\section{Limitations and Future Work}
\label{app:limitation}

Our framework inherits the common dependency of few-shot methods on the quality of normal reference images: if the selected references fail to cover the full variability of normal patterns, the resulting visual residuals may be contaminated by noise. Integrating robust reference selection strategies from the retrieval literature into our residual construction is a straightforward extension that we leave for future work. 

Additionally, the current residual branch treats the dot-product projection as a single scalar score, without explicitly decomposing the magnitude and angular components of the residual vectors, which capture complementary aspects of anomaly deviation. Deeper exploitation of this geometric structure could reduce the reliance on separate text and visual branches, yielding a more compact architecture with faster inference. We plan to explore this in future work.

\section{Licenses of Existing Assets}
\label{sec:asset_licenses}

In this work, we rigorously respect the licenses and terms of use of all existing datasets and foundation models. A detailed summary of the assets utilized, along with their respective public licenses, is provided in \cref{tab:asset_licenses}. 

\begin{table}[h]
  \caption{Summary of existing assets utilized in this work and their respective licenses.}
  \label{tab:asset_licenses}
  \centering
  \begin{tabular}{llcl}
    \toprule
    \textbf{Asset} & \textbf{Asset Type} & \textbf{Citation} & \textbf{License} \\
    \midrule
    CLIP (ViT-L/14) & Pre-trained Model & \cite{radford2021learning} & MIT License \\
    MVTec AD & Dataset & \cite{bergmann2019mvtec} & CC BY-NC-SA 4.0 \\
    VisA & Dataset & \cite{zou2022spot} & CC BY 4.0 \\
    BTAD & Dataset & \cite{mishra2021vt} & CC BY-SA 4.0 \\
    MPDD & Dataset & \cite{jezek2021deep} & CC BY-NC-SA 4.0 \\
    DTD-Synthetic & Dataset & \cite{aota2023zero} & Open Academic Use \\
    \bottomrule
  \end{tabular}
\end{table}





\end{document}